\def\preprintflag{1}

\documentclass{article}

\usepackage{microtype}
\usepackage{graphicx}
\usepackage{booktabs} 

\usepackage[T1]{fontenc}    
\usepackage{xr-hyper}
\ifdefined\preprintflag\else  
\ifdefined\appendixonlyflag
  \IfFileExists{main_mlsys.aux}{\externaldocument[][nocite]{main_mlsys}}{\typeout{No file main_mlsys.aux.}}
\else
  \IfFileExists{appendix_mlsys.aux}{\externaldocument[][nocite]{appendix_mlsys}}{\typeout{No file appendix_mlsys.aux.}}
\fi
\fi
\usepackage{hyperref}       
\usepackage{url}            
\usepackage{xurl}
\newcommand{\Capp}[1]{Appendix~\ref{#1}}
\ifdefined\appendixonlyflag
  \newcommand{\Cmain}[1]{\Cref{#1} of the main paper}
\else
  \newcommand{\Cmain}[1]{\Cref{#1}}
\fi
\makeatletter
\g@addto@macro\normalsize{%
  \setlength\abovedisplayskip{4pt plus 1pt minus 2pt}%
  \setlength\belowdisplayskip{4pt plus 1pt minus 2pt}%
  \setlength\abovedisplayshortskip{1pt plus 1pt}%
  \setlength\belowdisplayshortskip{2pt plus 1pt}}
\makeatother           
\usepackage{amsfonts}       
\usepackage{nicefrac}       
\usepackage{setspace} 
\usepackage{refcount} 
\usepackage{upquote}  
\usepackage{fancyhdr}
\usepackage{lastpage}       
\usepackage{lscape}         
\usepackage{verbatim}       
\usepackage{listings}       
\usepackage{epsfig}         
\usepackage{array}          
\usepackage{enumitem}       
\usepackage{hhline}         
\usepackage{siunitx}        
\usepackage{amsmath}        
\usepackage{amssymb}        
\usepackage{amsthm}         
\usepackage{ifthen}         
\usepackage{float}
\usepackage{cleveref}
\usepackage{booktabs}
\usepackage{makecell}

\usepackage{pdfpages}
\usepackage{subcaption}
\usepackage{multirow}
\usepackage{multicol}
\usepackage{wrapfig}
\usepackage[normalem]{ulem}
\usepackage{soul}
\useunder{\uline}{\ul}{}
\usepackage{pifont}
\newcommand{\cmark}{\textcolor{jcgreen}{\ding{51}}}%
\newcommand{\xmark}{\textcolor{jcred}{\ding{55}}}%
\usepackage{stmaryrd}
\newcommand{\sem}[1]{\llbracket #1 \rrbracket}
\usepackage{tabularx}

\usepackage{xcolor}

\definecolor{keywordcolor}{rgb}{0.7, 0.1, 0.1}   
\definecolor{tacticcolor}{rgb}{0.0, 0.1, 0.6}    
\definecolor{commentcolor}{rgb}{0.4, 0.4, 0.4}   
\definecolor{symbolcolor}{rgb}{0.0, 0.1, 0.6}    
\definecolor{sortcolor}{rgb}{0.1, 0.5, 0.1}      
\definecolor{diffdel}{rgb}{0.75, 0.1, 0.1}      
\definecolor{diffadd}{rgb}{0.1, 0.5, 0.1}       
\definecolor{jcredl}{HTML}{FB8072}              
\definecolor{jcgreenl}{HTML}{B3DE69}            
\definecolor{jcred}{HTML}{e31a1c}
\definecolor{jcgreen}{HTML}{33a02c}
\definecolor{attributecolor}{rgb}{0.7, 0.1, 0.1} 
\lstdefinelanguage{ptx}{
  morekeywords={.visible,.entry,.param,.reqntid,.u64,.u32,.b32,.b16,.f32,.f16,.pred,mov,shr,and,add,setp,bra,bar,ldmatrix,mma,cvt,st,ld,ret},
  morecomment=[l]{//}, sensitive=true, alsoletter={.}
}
\makeatletter
\newcommand{\difflinebg}[2][1]{\makebox[0pt][l]{\kern-3pt\color{#2}
  \smash{\rule[\dimexpr\ht\strutbox-#1\baselineskip\relax]{\dimexpr\lst@linewidth+6pt\relax}{#1\baselineskip}}}}
\makeatother
\lstdefinelanguage{torch}{
  morekeywords={class,def,return,import,from,self,super},
  morekeywords=[2]{torch,nn,Module},
  morecomment=[l]{\#},
  morestring=[b]", morestring=[b]',
  sensitive=true
}
\usepackage[most]{tcolorbox}
\newtheoremstyle{premisestyle}
  {0pt}{0pt}                  
  {\normalfont}{}
  {\bfseries\scshape}{.}{0.6em}
  {\thmname{#1}\thmnumber{ #2}\thmnote{ (#3)}}
\theoremstyle{premisestyle}

\tcolorboxenvironment{premise}{
  enhanced, breakable,
  colback=black!3, colframe=black!3,
  borderline west={1.5pt}{0pt}{green!50!black},
  boxrule=0pt, arc=0pt,
  left=6pt, right=6pt, top=4pt, bottom=4pt,
  before skip=8pt, after skip=8pt
}
\crefname{premise}{Workload Premise}{Workload Premises}
\Crefname{premise}{Workload Premise}{Workload Premises}

\tcolorboxenvironment{problem}{
  enhanced, breakable,
  colback=black!3, colframe=black!3,
  borderline west={1.5pt}{0pt}{black!60},
  boxrule=0pt, arc=0pt,
  left=6pt, right=6pt, top=4pt, bottom=4pt,
  before skip=8pt, after skip=8pt
}
\newtheorem{obligation}{Obligation}
\tcolorboxenvironment{obligation}{
  enhanced, breakable,
  colback=black!3, colframe=black!3,
  borderline west={1.5pt}{0pt}{green!50!black},   
  boxrule=0pt, arc=0pt,
  left=6pt, right=6pt, top=4pt, bottom=4pt,
  before skip=8pt, after skip=8pt
}
\crefname{obligation}{Obligation}{Obligations}
\Crefname{obligation}{Obligation}{Obligations}
\crefname{problem}{Problem}{Problems}
\Crefname{problem}{Problem}{Problems}

\tcolorboxenvironment{assumption}{
  enhanced, breakable,
  colback=black!3, colframe=black!3,
  borderline west={1.5pt}{0pt}{orange!70!black},   
  boxrule=0pt, arc=0pt,
  left=6pt, right=6pt, top=4pt, bottom=4pt,
  before skip=8pt, after skip=8pt
}
\crefname{assumption}{Modeling Assumption}{Modeling Assumptions}
\Crefname{assumption}{Modeling Assumption}{Modeling Assumptions}

\usepackage[scaled=0.85]{beramono}          
\definecolor{kw}{RGB}{0,72,150}
\definecolor{kw2}{RGB}{120,40,140}
\definecolor{cm}{RGB}{120,120,120}
\definecolor{str}{RGB}{160,60,20}

\newtcblisting{code}[2][]{
  listing only, listing engine=listings,
  listing options={language=#2},
  enhanced, breakable,
  colback=black!3, colframe=black!3,
  borderline west={1.5pt}{0pt}{black!60},
  boxrule=0pt, arc=0pt,
  left=6pt, right=4pt, top=3pt, bottom=3pt,
  before skip=0pt, after skip=0pt,
  #1
}

\ifdefined\preprintflag
  \usepackage[accepted]{mlsys2025}
  \makeatletter
  \renewcommand{\mlsys@appearing}{Preprint.}
  \makeatother
\else
\usepackage{mlsys2025}

\fi

\mlsystitlerunning{Lean-Assisted Reasoning for LLM-Generated GPU Kernels}

\begin{document}

\twocolumn[
\ifdefined\appendixonlyflag
\mlsystitle{Appendix: LARK: Lean-Assisted Reasoning for LLM-Generated GPU Kernels}
\else
\mlsystitle{LARK: Lean-Assisted Reasoning\\ for LLM-Generated GPU Kernels}
\fi



\mlsyssetsymbol{equal}{*}

\begin{mlsysauthorlist}
\mlsysauthor{Zixi Zhang}{ic}
\mlsysauthor{Jiayi Nie}{cam}
\mlsysauthor{Haoran Wu}{cam,qu}
\mlsysauthor{Yujie Yan}{ed}
\mlsysauthor{Yann Herklotz}{ef}
\mlsysauthor{Rika Antonova}{cam}
\mlsysauthor{Robert Mullins}{cam}
\mlsysauthor{Yiren Zhao}{ic}
\mlsysauthor{George A. Constantinides}{ic}
\mlsysauthor{Jianyi Cheng}{ed}
\end{mlsysauthorlist}

\mlsysaffiliation{ic}{Department of Electrical and Electronic Engineering, Imperial College London, London, United Kingdom}
\mlsysaffiliation{cam}{Department of Computer Science and Technology, University of Cambridge, Cambridge, United Kingdom}
\mlsysaffiliation{ed}{School of Informatics, University of Edinburgh, Edinburgh, United Kingdom}
\mlsysaffiliation{ef}{Computer Science Department, EPFL, Lausanne, Switzerland}
\mlsysaffiliation{qu}{Quettaflop AI}

\mlsyscorrespondingauthor{George A. Constantinides}{g.constantinides@imperial.ac.uk}
\mlsyscorrespondingauthor{Jianyi Cheng}{jianyi.cheng@ed.ac.uk}


\vskip 0.3in

\ifdefined\appendixonlyflag\else
\begin{abstract}
Efficient GPU kernels are typically low-level and hardware-specific, and LLM agents can now generate them faster than experts can verify.
Existing approaches rely on testing, which provides evidence but not a guarantee of correctness.
We introduce \textbf{LARK}, an agentic framework that automatically and formally verifies GPU kernels against their PyTorch reference implementations.
LARK proves equivalence in functional and memory behavior under exact arithmetic and establishes a new \emph{trust boundary}: human effort is only needed to verify operator-level Lean semantics ahead of kernel generation time, instead of checking every single line of generated code.
LARK formalizes the GPU machine model and the semantics of each operation in Lean 4 once per target language, and the Lean 4 semantics of any incoming kernel is derived from these pre-built definitions.
LLM agents then automate the proof in two steps.
First, an agent decomposes the whole kernel semantics into smaller segments and Lean 4 verifies this segmentation, which addresses the intractability of proving equivalence for large kernels directly.
Second, agents write Lean proofs that each kernel segment matches its high-level PyTorch counterpart. Together, the two steps give whole-program equivalence.
We demonstrate that LARK can be used 1) to ensure correctness in existing code-generation agents and 2) to automatically verify human-written kernel libraries. On the KernelCraft GPU tasks, combined with a simple ReAct agent, LARK enables GLM 5.3 to fully verify 10 of 12 Level-1 kernels and Opus 5.5 all 5 Level-2 kernels.
Moreover, over 10 production kernels randomly sampled from popular open-source LLM serving frameworks, LARK verifies 9 and identifies a new data race bug in the tenth.
We believe LARK represents a significant step toward verifiable code generation in the agentic era.
\end{abstract}
\fi  
]



\printAffiliationsAndNotice{}  

\ifdefined\appendixonlyflag\else
\section{Introduction}
\label{sec:intro}

\begin{figure*}
    \centering
    \includegraphics[width=\linewidth]{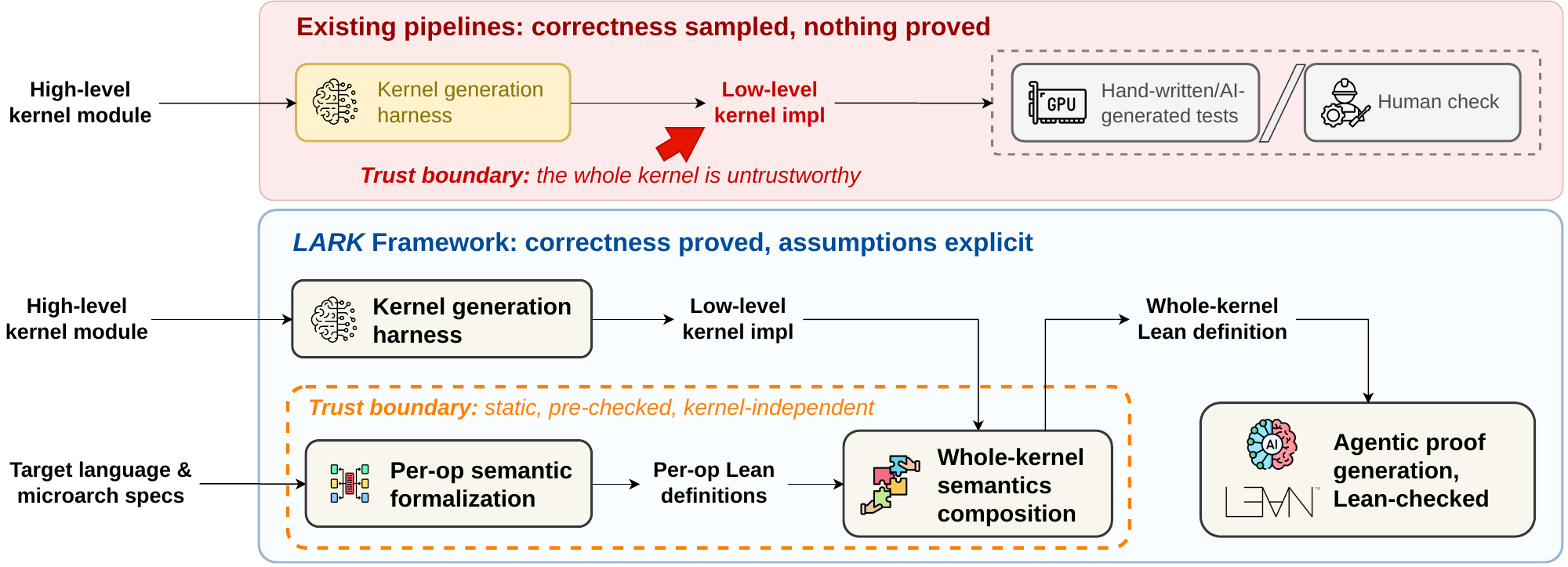}
    \caption{Where trust sits in LLM kernel generation. \textbf{Top:} existing pipelines check a generated kernel by sampled tests or review, so the whole kernel is taken on faith. \textbf{Bottom:} LARK fixes what is assumed (dashed) before the kernel is seen, the machine model and operation semantics formalized once per language and the task's interface; every property of the kernel is proved in Lean or the kernel is rejected, and the assumptions are reported with every result.}
    \label{fig:flowchart-compare}
    \vspace{-1em}
\end{figure*}

GPU kernels are where the cost of training and serving deep learning models is paid, and the fastest are hand-scheduled CUDA, PTX, or HIP written close to the metal~\citep{cutlass,flashattn,triton}. Such code is increasingly written by LLM agents~\citep{kernelcraft,kernelbench,sakanaai,kevin}, and a plausible kernel is not a correct one: the same models that write fluent code write fluent bugs~\citep{lange2025robustagenticcudakernel,proofwright}. Every regenerated kernel needs its own correctness check, at the pace of generation.
Existing checks fall short on one count or the other (\Cref{fig:flowchart-compare}, top). Tests on sampled inputs keep pace but give evidence, never a guarantee; human review reasons about all inputs but cannot keep pace; an LLM judge removes the human but not the problem. What a generated kernel needs is a proof of a precise equivalence statement, checked by a proof assistant such as Lean~4~\citep{lean4}, and formal tools do not provide one either: GPUVerify~\citep{gpuverify} proves race and divergence freedom, not the specification; a verified compiler~\citep{compcert} certifies its own translation, not a generated kernel; a hand-written model~\citep{sail} or proof~\citep{sel4} puts a human back where we want automation.

We present \textbf{LARK} (Lean-Assisted Reasoning for Kernels), an agentic framework in which LLM agents write both the model of the kernel and the proof, and Lean checks both. Once per target language, LARK formalizes in Lean a machine model of the GPU and the semantics of each operation and measures them against the hardware; the machine model is shared across languages, so a new target supplies only its operations and their semantics, and we instantiate LARK for PTX, CUDA, and AMD HIP. Given a PyTorch specification and a kernel from any generator, the kernel's semantics is derived from these pinned parts, agents write the proofs, and Lean checks every step, so a wrong agent output is a failed proof, never a wrong theorem. This division of labor raises three challenges:

\textbf{1) Semantic gap.} The kernel's Lean semantics is spelled out operation by operation over the machine model, while the PyTorch reference has a tensor-level semantics in Lean. LARK bridges the two by deriving a \emph{flat model} of the kernel and having an LLM agent segment it, verifiably, into a \emph{segment-level model} that a proof can induct on (\Cref{sec:bridge}).

\textbf{2) Fidelity versus tractability.} The machine model must be sound for the hardware yet abstract enough to prove against. LARK's model computes over exact numbers, runs a phase's threads in sequence, and treats undefined behavior as a fault; what makes this sound, race freedom above all, is proved in Lean per kernel as \emph{obligations}, and a kernel that can fault or race has no theorem (\Cref{sec:foundations,sec:advanced}).

\textbf{3) Trust boundaries.} Any assumption the theorem depends on but Lean does not verify must be trusted by humans; we call this set the \emph{trust boundary}. LARK keeps it small and fixed by the task, the hardware machine model, the fixed mapping from operation semantics to kernel semantics, and the formalized task specification, so it can be reviewed once in advance; every run produces a Lean certificate that any Lean project re-checks without repeating the LARK procedure (\Cref{sec:trust}).

LARK tackles these three challenges, and we make the following contributions:

\begin{itemize}[leftmargin=1em, itemsep=0em, topsep=0em]
    \item We present LARK, an agentic formal-verification workflow for GPU kernels in low-level, close-to-metal languages. To our knowledge, LARK is the first attempt at using agents to produce formally verified low-level GPU code for equivalence check.
    \item We show how to formalize low-level GPU semantics in Lean 4 and prove functional equivalence between a GPU kernel and a PyTorch reference. LARK combines verified program segmentation with iterative agentic proof construction, and humans review the task's trust boundary \textbf{only once, ahead of time} in dependent of kernel inputs; every kernel generated afterwards is certified relative to those trusted assumptions.
    \item We demonstrate that LARK can be integrated with agentic code-generation harnesses to certify their kernels and serves as an automated verifier for human-written kernels. On KernelCraft, GLM~5.3 certifies 10 of the 12 Level-1 kernels and Opus~5.5 all five fused Level-2 kernels, every semantics-breaking mutant is rejected, and nine of ten production kernels verify while the tenth is rejected for a real data race (\Cref{sec:experiments}).
\end{itemize}

\section{Motivating Example}\label{sec:example}

\begin{figure*}
  \centering
  \begin{minipage}[t]{0.45\linewidth}
    \begin{subfigure}[t]{\linewidth}
      \begin{lstlisting}[language=torch, basicstyle=\fontsize{7}{7}\selectfont\ttfamily, escapeinside={(*@}{@*)}]
def matmul(A: torch.Tensor, B: torch.Tensor) -> torch.Tensor:
    assert A.shape == (64, 256)
    assert B.shape == (256, 512)
    assert A.dtype == B.dtype == torch.float16
    acc = A.float() @ B.float()
    return acc.half()
\end{lstlisting}
      \caption{Tensor-level Torch specification $S$.}
      \label{fig:running:torch}
    \end{subfigure}

    \vspace{8pt}
    \begin{subfigure}[t]{\linewidth}
      \begin{lstlisting}[language=lean, basicstyle=\fontsize{7}{7}\selectfont\ttfamily, escapeinside={(*@}{@*)}, deletekeywords={[1]def,theorem}, keywordstyle={[5]\ttfamily\bfseries\color{keywordcolor}}, morekeywords={[5]def,theorem}]
def gemmSig : Sig := [        -- the task's signature: role, type, shape, layout
  .buf .in  .f16 [64, 256] (rowMajor 256),   -- A
  .buf .in  .f16 [256, 512] (rowMajor 512),  -- B
  .buf .out .f16 [64, 512] (rowMajor 512) ]  -- C
def gemmLaunch : Launch := .pinned 32 1 32        -- 32 blocks of 32 threads

theorem kernel_eq_spec (s : MachineState) (A B)
   (h : StdPre gemmSig gemmLaunch s (A, B)) :       -- placed, separated, aligned, geometry
   (*@$\exists$@*) s', runLaunch (prog PTX gemm_src fuels) s = some s'
     (*@$\wedge$@*) stdExtract gemmSig s' = Spec.highlevel_spec A B := by
   ...   -- proof generated by LARK, Lean checked
\end{lstlisting}
      \caption{Equivalence theorem in Lean.}
      \label{fig:running:thm}
    \end{subfigure}
  \end{minipage}\hfill
  \begin{subfigure}[t]{0.53\linewidth}
    \begin{lstlisting}[language=ptx, numbers=left, numberstyle=\tiny\color{gray}, numbersep=4pt, xleftmargin=12pt, escapeinside={(*@}{@*)}, moredelim={*[il][\color{diffdel}]{@@-}}, moredelim={*[il][\color{diffadd}]{@@+}}]
.entry matmul_f16_mma(.param .u64 p_A, .param .u64 p_B, .param .u64 p_C)
.reqntid 32, 1, 1
{ mov.u32 %r0, %ctaid.x;  // output tile from block index
  ...    // load p_A, p_B, p_C; per-lane addresses from %tid.x; zero %f0..%f31
K_LOOP:  // 16 slabs of k = 16
  ld.global.b32 %w0, [%rd11];   ...   st.shared.b32 [%r30], %w0;  ...  // stage A, B slabs
  bar.sync 0;
  ldmatrix.sync.aligned.m8n8.x4.shared.b16       {%r30..%r33}, [%r24];  // A fragment
  ldmatrix.sync.aligned.m8n8.x4.trans.shared.b16 {%r34..%r37}, [%r25];  // B fragments
  mma.sync...m16n8k16.f32.f16.f16.f32 {%f0..%f3}, {%r30..%r33}, {%r34,%r35}, {%f0..%f3};
  ...    // 7 more mma (3 more ldmatrix) for n8 tiles 1..7
(*@\difflinebg[2]{jcredl!20}@*)@@-- // Bug: no barrier before it is re-staged
(*@\difflinebg[2]{jcredl!20}@*)@@-- @%p1 bra K_LOOP;
(*@\difflinebg[2]{jcgreenl!20}@*)@@++ // Fixed by LARK: every lane is done reading the slab
(*@\difflinebg[1]{jcgreenl!20}@*)@@++ bar.sync 0;  @%p1 bra K_LOOP; 
  cvt.rn.f16.f32 %h0, %f0;  st.global.b16 [%rd15], %h0;   // C[row0+g, col0+2t]
  ...  ret; }   // after 31 more cvt + st.global.b16 of C
\end{lstlisting}
    \caption{PTX kernel $K$, condensed from about 300 lines.}
    \label{fig:running:ptx}
  \end{subfigure}

  \caption{Running example: general matrix multiplication (GEMM).
    \textbf{(a)}~The PyTorch reference.
    \textbf{(b)}~The equivalence theorem: the task's fixed premise (\texttt{StdPre}: inputs placed, buffers apart, bases aligned, launch geometry) implies fault-free completion (\texttt{some s'}) and $AB$ in the region of $C$, over the reals.
    \textbf{(c)}~The PTX kernel implementing (a); its per-lane addressing, shared-memory staging, barrier, and warp-collective \texttt{ldmatrix}/\texttt{mma} (lines~4--11) each need formal meaning before (b) can be stated. Line~13 (red), in place of line~15 (green), omits the barrier before the next slab is staged.
    Testing can only sample (b) and accepts the bug; LARK proves (b) with line~15 and locates the race with line~13.}
  \label{fig:running}
  \vspace{-1em}
\end{figure*}

\Cref{fig:running} makes the three challenges of \Cref{sec:intro} concrete. The specification (\Cref{fig:running:torch}) multiplies two FP16 matrices. The kernel (\Cref{fig:running:ptx}) is LLM-written PTX, the virtual instruction set CUDA compiles to, in the shape of fast kernels: each one-warp block computes one output tile (line~3), staging slabs of $A$ and $B$ into shared memory (lines~5--7) and multiplying them on the tensor cores (lines~8--11) before storing the tile (lines~16--17). The theorem (\Cref{fig:running:thm}) relates the two through the state of a machine that runs the kernel, under a premise of the task, not of the kernel; a test samples it, a proof establishes it for every input (\Cref{sec:problem}).

\textbf{What a guarantee catches.} Line~13 of \Cref{fig:running:ptx} is a one-line bug a generator could write: without the barrier, a lane may overwrite a staged slab another lane is still reading. Nothing marks the omission among 300 lines that look alike, and the kernel can pass every test, since on a given GPU the warp may finish \texttt{ldmatrix} before any lane stores, while PTX promises no such order. LARK catches it in Lean: the race-freedom obligation it proves before any value (\Cref{sec:advanced}) fails for line~13 with a counterexample, two lanes meeting at one shared cell, that points to where the barrier belongs; with line~15 the kernel gets the theorem in \Cref{fig:running:thm}.

\textbf{Where the challenges show.} \textit{Semantic gap:} the conclusion speaks of the matrix $AB$, while the kernel updates four registers per lane per \texttt{mma} (line~10) and loops by a branch to a label (line~15), naming no tile, iteration, or sum to induct on. \textit{Fidelity versus tractability:} a model faithful to every interleaving of 32 lanes and 32 blocks admits no tractable proof, and each abstraction that helps has a condition, such as sequential threads needing lines~6 and~8 never to race, that must be proved rather than assumed for the theorem to be worth anything. \textit{Trust:} that the Lean \texttt{mma} is what the tensor core computes, that \texttt{bar.sync} is a barrier, and that the Lean specification and interface are those of the task can only be believed, so this trust boundary (\Cref{sec:intro}) must exclude whatever an agent writes and stay a short, fixed list; \Cref{sec:method} shows how LARK meets each challenge.

\section{Problem Definition}
\label{sec:problem}

\Cref{sec:example} showed the theorem informally; we now state, for a fixed target language, what is proved, what is assumed, and what a verifier must deliver.

\textbf{Specification.}
A \emph{specification} $S$ is a Torch module, denoting a total function $\sem{S} : D \to T$ from the inputs $D$ with the declared shapes, types, and constraints to the output tensors $T$. $S$ is the ground truth.

\textbf{Kernel and machine model.}
A \emph{kernel} $K$ is a program in the target language with a launch geometry. Its meaning is fixed only relative to a \emph{machine model} $M$: the states $\Sigma_M$, a partial semantics for each operation, a scheduling rule, and a representation of values (ours is in \Cref{sec:foundations}). These define the partial \emph{whole-launch semantics} $\sem{K}_M : \Sigma_M \rightharpoonup \Sigma_M$, defined at $\sigma$ when every maximal execution from $\sigma$ is finite, fault-free, and ends in the same state.

\textbf{Interface.}
Tensors enter and leave the machine through memory under the task's \emph{interface} $I$: the region and layout of each tensor, the launch geometry, and the disjointness of the regions. $I$ belongs to the task, not to $K$, and yields a \emph{precondition} $\mathrm{In}_I(\sigma, x)$, true when $\sigma$ has the declared geometry and holds $x \in D$ in the declared regions, and an \emph{output extractor} $\mathrm{Out}_I : \Sigma_M \to T$. In \Cref{fig:running:thm}, \texttt{StdPre} is $\mathrm{In}_I$, \texttt{stdExtract} is $\mathrm{Out}_I$, and \texttt{runLaunch} is $\sem{K}_M$.

\textbf{Equivalence.}
Both sides are compared as real-valued tensors, the kernel's operations read over exact reals (\Cref{sec:foundations,sec:trust}). $K$ is \emph{equivalent to $S$ under $M$ and $I$}, written $K \equiv_{M,I} S$, if
\begin{equation}\label{eq:equiv}
\begin{aligned}
  &\forall x \in D.\;\forall \sigma \in \Sigma_M.\quad
  \mathrm{In}_I(\sigma, x) \\
  & \Longrightarrow\; \sem{K}_M(\sigma){\downarrow}
  \;\wedge\; \mathrm{Out}_I\big(\sem{K}_M(\sigma)\big) = \sem{S}(x).
\end{aligned}
\end{equation}
Quantifying over every initial state that holds the input forbids $K$ from depending on memory outside the interface, and reading the output from the declared region fixes where the result lands. This is all that equivalence in \emph{memory behavior} means: the memories of $M$ are unbounded (\Cref{sec:foundations}), so a write outside the interface is invisible to the statement, and out-of-bounds safety on the host's allocations is not guaranteed.

\textbf{Trust boundary.}
A proof of \Cref{eq:equiv} is a proof about $M$; to carry over to the real hardware and host $\mathcal{H}$, it needs facts no proof inside $M$ can establish: that $M$ is faithful to the hardware up to its idealizations, that the formalization of $S$ denotes $\sem{S}$, and that the host honors $I$. This trust boundary must be \emph{fixed by the task, not by the kernel}: $M$ and $I$ are unavoidably premises, nothing else may be, and properties of $K$ such as race freedom or termination are \emph{obligations}, proved per kernel (\Cref{sec:foundations,sec:advanced}).

\textbf{Problem definition.}
Fix a target language with machine model $M$. Given a specification $S$, an interface $I$, and a kernel $K$, produce either a \emph{certificate} $(\Pi, \pi)$ or a \emph{rejection}, which makes no claim about $K$. The premises $\Pi = \Pi_M \cup \Pi_K$ are $\Pi_M$, the faithfulness of $M$ to $\mathcal{H}$, and $\Pi_K$, the task's signature and launch, from which $\mathrm{In}_I$ and $\mathrm{Out}_I$ are computed by fixed definitions; both are reported with the result (\Cref{sec:trust}). The machine-checked proof $\pi$ establishes
\[
  \Pi_K \;\vdash\; K \equiv_{M,I} S ,
\]
with $\mathrm{In}_I$ satisfiable for every input in $D$, and hence $\Pi_M \models K \equiv_{\mathcal{H},I} S$: $\pi$ supplies the equivalence in $M$, $\Pi_M$ the passage from $M$ to $\mathcal{H}$.

\textbf{Requirements.}
A solution must be \emph{sound}: a certificate is issued only if $K \equiv_{M,I} S$ holds, and $\pi$ is checked by a checker that does not trust its producer. Soundness makes both conjuncts of \Cref{eq:equiv} obligations, proved rather than assumed. Among sound solutions we prefer the \emph{more complete}, certifying more of the kernels that are equivalent, and the \emph{more trustworthy}, certifying with a smaller $\Pi_M$. The second objective separates this problem from classical equivalence checking: race freedom and the other regularities that make GPU kernels tractable are properties of $K$, proved per kernel, so that $\Pi_M$ keeps only what no per-kernel proof can discharge.

\section{Method}\label{sec:method}

\begin{figure*}
    \centering
    \includegraphics[width=\linewidth]{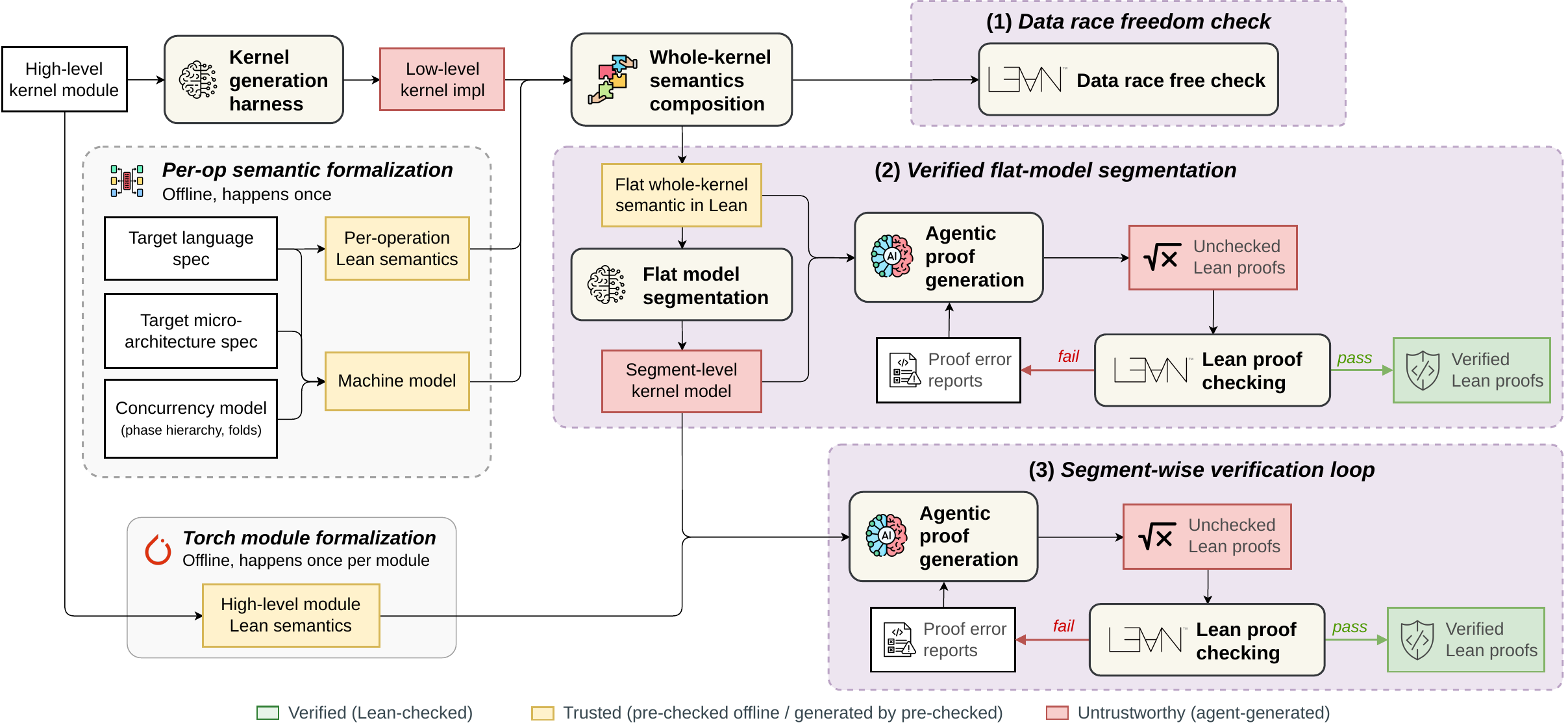}
    \caption{The LARK pipeline. Offline, the machine model and per-operation semantics are formalized once per target language and measured against the hardware, and the Torch module once per module. Per kernel, the translator composes the flat whole-kernel semantics in Lean, then three Lean-checked stages follow: (1)~the race-freedom check; (2)~an agent segments the flat model and proves the segment-level model equivalent to it; (3)~agents prove the segment-level model equivalent to the Torch semantics. Lean's error reports drive the agent loops. Green: Lean-checked. Yellow: trusted, listed in the report (\Cref{sec:trust}). Red: agent-generated, trusted for nothing.}
    \label{fig:flowchart-detail}
    \vspace{-1em}
\end{figure*}

\Cref{sec:problem} asks for a certificate whose premises say nothing about the kernel: everything inside the trust boundary must be fixed before the kernel arrives, and everything about the kernel must be proved. This section follows that split (\Cref{fig:flowchart-detail}): a Lean model of the kernel we can trust, fixed once per target language (\Cref{sec:foundations}); the way from the model to a proof, written by agents and checked by Lean (\Cref{sec:bridge,sec:advanced}); and the short, fixed list of what is left to trust (\Cref{sec:trust}).

\subsection{A Model We Can Trust}\label{sec:foundations}

In this section, we describe how, \textbf{ahead of time}, a Lean semantic model can be generated for each operation in a target language such as PTX or HIP. This process involves three key components:
1) Machine model: Defining a machine model encompassing memory, register abstraction, fault handling, and concurrency, 
2) Per-op semantic contextualization: Specifying per-operation semantics that contextualize the machine model for the chosen target language; 
and 3) Translator: Developing translators to automate kernel semantic composition and support extension to new target languages.

In this setup, the Lean model of a kernel can be trusted because everything it is built from, a machine model, the semantics of each operation, and a translator, is fixed and checked by human experts before the kernel arrives, and none of it is generated by agents.

\textbf{Machine model.} The machine model of \Cref{sec:problem} is a quadruple $M = (\Sigma, \mathcal{I}, \mathrm{sem}, \mathrm{cut})$: the states, the operations of the target language, their semantics, and their classification by the synchronization they perform. Each part trades fidelity for tractability, once, for every kernel.

\textit{Memory.} A state holds the whole launch:
\[
  \Sigma = \mathit{pc} \times \mathit{reg} \times \mathit{mem}_{\mathrm g} \times \mathit{mem}_{\mathrm s} \times \mathit{mem}_{\mathrm l} \times \mathit{geom},
\]
with $\mathit{pc}$ the program counters, $\mathit{reg}$ the registers, keyed by thread and name, $\mathit{mem}_{\mathrm g}$ the global memory, $\mathit{mem}_{\mathrm s}$ one shared memory per block, $\mathit{mem}_{\mathrm l}$ one local memory per thread, and $\mathit{geom}$ the read-only launch geometry and coordinates. Keeping every thread's registers in one state makes it immediate that an operation of one thread touches no register of another. Memories are unbounded, so the model cannot detect an access outside the host's allocations, which is outside the guarantee (\Cref{sec:problem}).\label{asm:host}

\textit{Values.} Every register and memory cell holds a typed value or is undefined:
\[
  \mathit{Val} = \{\bot\} \cup \{(\tau, x) : \tau \in \mathit{Ty},\ x \in \mathit{Payload}(\tau)\},
\]
a type tag $\tau \in \mathit{Ty}$ (the set is listed in \Capp{app:values}) and a payload $x$: an exact real, extended with $-\infty$ for running maxima, under a float tag, an integer under an integer tag, a bit under a predicate. Exact arithmetic is a deliberate idealization, stated once as the scope of the guarantee rather than as an assumption about any kernel: LARK does not track numerical error, \Cref{eq:equiv} is equivalence under exact arithmetic,\label{asm:real} and floating-point behavior is left to testing checks (\Cref{sec:trust}).

\textit{Faults.} $\mathrm{sem}_t(i) : \Sigma \rightharpoonup \Sigma$ is the semantics of operation $i$ executed by thread $t$, which selects the register row and the coordinates. Its undefined cases are the \emph{faults}: an IEEE result outside $\mathbb{R} \cup \{-\infty\}$, a read of the wrong tag or of $\bot$, a misaligned vector access, a negative address, an exhausted fuel (defined below), a non-converged cut. A kernel that can fault has no theorem.

\textit{Concurrency.} Reasoning about every interleaving of thousands of threads is not practical, and machine-learning kernels do not need it: they work in \emph{phases} separated by barriers, within which no two threads touch the same data. The model makes phases explicit through $\mathrm{cut} : \mathcal{I} \to \{\bot, \mathrm{coll}, \mathrm{warp}, \mathrm{block}, \mathrm{grid}, \mathrm{halt}\}$ (\Cref{tab:instr-cats}): a barrier is a cut at its level, a return the halt, and a warp-collective operation such as \texttt{mma} a cut of kind $\mathrm{coll}$, where the lanes of the warp meet and the operation executes once over all their registers. A kernel is a sequence $K \in \mathcal{I}^*$ with $K[p]$ the operation at program counter $p$; a thread iterates $\mathrm{sem}$ and \emph{parks} at a cut:
\[
  \mathrm{step}_{K,t}(\sigma) =
  \begin{cases}
    \mathrm{sem}_t(K[p])(\sigma) & \text{if } \mathrm{cut}(K[p]) = \bot, \\
    \sigma & \text{otherwise (parked)},
  \end{cases}
\]
where $p = \mathit{pc}_t(\sigma)$. A launch is a fold over the hierarchy of units, in time and then in parallelism:
\begin{multline}\label{eq:fold}
  \mathrm{run}_{\ell}(\sigma) = \\
  \mathrm{conv}_{\ell}\big(\mathrm{run}_{\ell-1}(u_m) \circ \cdots \circ \mathrm{run}_{\ell-1}(u_1)\big)(\sigma).
\end{multline}
Here $u_1, \dots, u_m$ are the units one level down (the lanes of a warp, the warps of a block, the blocks of the grid) making up the unit at level $\ell \in \{\mathrm{warp}, \mathrm{block}, \mathrm{grid}\}$, each run to the next cut of level $\ell$; $\mathrm{conv}_\ell(f)(\sigma) = f(\sigma)$ if every thread of the unit parks at one program counter in $f(\sigma)$, undefined otherwise; and $\mathrm{run}_0$ iterates $\mathrm{step}$ to the lane's next cut. Each level has a \emph{fuel} $\kappa_\ell \in \mathbb{N}$, the steps or sub-runs allowed before its next cut; $\mathrm{run}^{0}_\ell$ is undefined, so running out of fuel is a fault, and $\vec\kappa$ is data the run supplies. The theorem requires $\sigma \in \mathrm{dom}(\mathrm{run}^{\vec\kappa}_{\mathrm{grid}})$.

\textbf{Per-operation semantics.} Each operation $\mathrm{op}$ has a lower-layer function $\widehat{\mathrm{op}}$, plain mathematics independent of the machine ($\widehat{\mathrm{add}} = {+}$, $\widehat{\mathrm{max}} = \max$ on $\mathbb{R} \cup \{-\infty\}$, $\widehat{\mathrm{mma}}$ a tile multiply-accumulate), and an upper layer that binds it to the state. Writing an operation as in the kernel, $\mathrm{op}\ d, a_1, \dots, a_k$, with register names $d$ and $a_j$, and $\tau_j$ and $\tau$ for the tags the operation names for operand $j$ and the result, a scalar operation executed by thread $t$ is
\[
\begin{aligned}
  &\mathrm{sem}_t(\mathrm{op}\ d, a_1, \dots, a_k)(\sigma) \\
  &\quad= \sigma\big[\mathit{reg}(t,d) \mapsto (\tau, \widehat{\mathrm{op}}(x_1, \dots, x_k)),\ \mathit{pc}_t \mapsto \mathit{pc}_t + 1\big]
\end{aligned}
\]
if $\mathit{reg}(t, a_j) = (\tau_j, x_j)$ for every $j$ and $(x_1, \dots, x_k) \in \mathrm{dom}(\widehat{\mathrm{op}})$, and undefined otherwise. \Cref{tab:instr-cats} gives the schema of each kind of operation: the guard that bounds the domain, the update, and the cut. The operations of each language and their lower-layer definitions are generated from one pinned table per language (\Cref{fig:flowchart-detail}, per-op semantic formalization), and every definition is measured against a real GPU on random operands before use (\Capp{app:non-ctx-sem-fuzzing}).

\begin{table}
  \centering
  \setlength{\tabcolsep}{3pt}
  \resizebox{\columnwidth}{!}{%
  \begin{tabular}{@{}p{2cm}llp{1.8cm}@{}}
    \toprule
    Kind\newline (examples) & Guard & Update & $\mathrm{cut}$ \\
    \midrule
    Scalar\newline (\texttt{add}, \texttt{mul}) & tags, $\vec x \in \mathrm{dom}\,\widehat{\mathrm{op}}$ & $\mathit{reg}(t,d) \mapsto (\tau, \widehat{\mathrm{op}}(\vec x))$ & $\bot$ \\
    Memory\newline (\texttt{ld}, \texttt{st}) & tag, aligned, $\alpha \ge 0$ & $\mathit{reg}(t,d) \leftrightarrow \mathit{mem}(\alpha)$ & $\bot$ \\
    Control\newline (\texttt{bra}, \texttt{ret}) & $\mathit{reg}(t,p) = (\mathrm{pred}, b)$ & $\mathit{pc}_t \mapsto q$ if $b$ & $\bot$, halt \\
    Barrier\newline (\texttt{bar.sync}) & --- & parks & warp, block, grid \\
    Collective\newline (\texttt{mma}) & tags on all lanes & all lanes' $\mathit{reg}(t',d)$ & coll \\
    \bottomrule
  \end{tabular}}
  \caption{Operation schemas by kind: guard (failed means a fault), update, and cut; all but Control also advance $\mathit{pc}_t$. $\widehat{\mathrm{op}}$ is the operation's own function ($+$, an address $\alpha$, a branch target $q$, a tile product); $\mathrm{sem}$ is defined per operation, the kinds only group them.}
  \label{tab:instr-cats}
  \vspace{-2em}
\end{table}

\textbf{Translator.} The translator is fixed code, not an agent, and it never guesses. It reads the kernel text $b$ as data, $\mathrm{parse}_L : \mathit{Bytes} \rightharpoonup \mathcal{I}^*$ (CUDA and HIP through a structured tree whose lowering $\mathrm{lower}_L : \mathit{Tree} \to \mathcal{I}^*$ is a definition in $M$), so the kernel's fold is $\mathrm{run}^{\vec\kappa}_{\mathrm{grid}}$ on $K = \mathrm{parse}_L(b)$ and nothing of the semantics is generated per kernel. Every reading is checked by a \emph{read-back}: a formatter $\mathrm{fmt}_L : \mathcal{I}^* \to \mathit{Bytes}$ writes the data back and $\mathrm{fmt}_L(\mathrm{parse}_L(b)) = b$ must hold byte for byte, so what is trusted is the formatter, a quarter the size of the reader (\Capp{app:readback}). An operation or pattern outside $\mathcal{I}$ or the concurrency model (asynchronous operations, atomics, a NaN literal) is refused by name before any proof is attempted, since we would rather refuse a correct kernel than prove a wrong theorem (\Capp{app:unsupported}). Finally, the translator prepares the proof with \emph{proved lemmas}, generated with their proofs and checked by Lean (the race-freedom lemmas of \Cref{sec:advanced}, the lane bridges and program-counter rungs of \Cref{eq:fold}), and \emph{scaffolding lemmas}, whose statements it generates and whose proofs an agent writes, so they can be false and then simply fail. Two generator hints, the exact constant behind a rounded literal and the expected launch, guide this generation but decide nothing; Lean can check neither, so both are reported as trusted (\Cref{sec:trust}).

\textbf{Retargeting.}\label{sec:retarget} $M$ splits as $M_{\mathrm{launch}} \cup L$: a launch part shared by every language and a per-language part $L$, which includes its operations with their semantics and cuts, warp width, reader and read-back formatter, collective lane layouts, and lowering (\Capp{app:retarget}). We instantiate LARK for PTX, CUDA, and HIP.

\subsection{From Model to Proof}\label{sec:bridge}

The automated Lean proof flow relies on two key checks: ensuring data race freedom and verifying functional equivalence. Data race freedom can be directly verified within Lean (\Cref{fig:flowchart-detail} Stage 1); on the other hand, verifying full functional equivalence may pose computational challenges. To address this, we propose a segment-level approach: first, perform a formally verified flat-model segmentation (\Cref{fig:flowchart-detail} Stage 2); then, check functional equivalence at the segment level (\Cref{fig:flowchart-detail} Stage 3).

\textbf{Stage 1: Data-race-freedom check.}\label{sec:advanced} The fold of \Cref{eq:fold} runs the units of a phase one after another, while the hardware runs them at once. The two agree only for race-free kernels, so race freedom is the first thing LARK proves about a kernel, before any equivalence is attempted. The vocabulary is the fold's: a \emph{phase} $\phi$ at a level is the run between two consecutive cuts of that level, its \emph{units} $u$ are the lanes, warps, or blocks that the level folds, and the footprints $W_\phi(u), R_\phi(u) \subseteq \mathit{Loc}$ are the locations unit $u$ writes and reads during $\phi$, with $\mathit{Loc}$ the global, shared, and local addresses and the register names of $\Sigma$.

\begin{obligation}[Race freedom]\label{prem:indep}
$\mathrm{RF}(K)$: for every phase $\phi$ and every two distinct units $u_j, u_k$ of it,
$W_\phi(u_j) \cap W_\phi(u_k) = \emptyset$ and $W_\phi(u_j) \cap R_\phi(u_k) = \emptyset$.
\end{obligation}
A warp-collective step counts as one more unit: it executes once per warp, with a footprint over the registers of every lane and the memory it reads, and it must meet no lane's footprint within the warp-level phase around its $\mathrm{coll}$ cut, since that cut orders execution, not memory. Race freedom is what the fold needs: the actions of race-free units commute and the barrier that ends a phase orders the phases, so every execution from $\sigma$ is finite and ends in the fold's result, $\sem{K}_M(\sigma) = \mathrm{run}^{\vec\kappa}_{\mathrm{grid}}(\sigma)$. This is Lipton's reduction~\citep{parallelproveproperty} as it applies to data-race-free GPU kernels~\citep{weakordering,InterleavingAL,gpuverify,formalptxmemconsist}; we rely on it at every level as metatheory, not as a Lean proof (\Cref{sec:trust}). To discharge the obligation, the translator derives each footprint from the kernel's addresses and generates the two lemmas of \Cref{prem:indep}, with proofs, for every phase and pair of units (\Capp{app:racefree}); Lean checks them, an agent attempts any lemma whose proof fails, and a kernel with an unproved lemma is rejected. The condition is sufficient, not necessary: a benign race is rejected as readily as a harmful one (\Cref{sec:experiments} reports one found in a production kernel), the price of never accepting a racy kernel.

\textbf{Stage 2: Verified flat-model segmentation.} The kernel's \emph{flat model} $F_K = \mathrm{run}^{\vec\kappa}_{\mathrm{grid}}$ on $K$ follows the kernel line by line with a program counter. Built only from the pre-checked parts of \Cref{sec:foundations} and the kernel's text, it is trustworthy, and it is one semantics for every language, so CUDA and HIP are lowered to it at no cost in trust. It is not the proof's view: it talks about single operations and registers where the Torch module talks about tensors, and a direct proof would unfold it operation by operation with no index to induct on. An agent therefore segments the flat model into a \emph{segment-level model} $G_K$, a term of the grammar
\[
\begin{aligned}
  G ::={}& \mathrm{sem}_t(i) \mid G ; G \mid \mathrm{for}\ j < n\ \mathrm{do}\ G(j) \\
  & \mid \mathrm{if}\ p\ \mathrm{then}\ G\ \mathrm{else}\ G,
\end{aligned}
\]
read as a partial function on states like $F_K$: runs of operations become a sequence, repeated runs one body $G(j)$ under an index, and a loop a $\mathrm{for}$ over its body. For the GEMM of \Cref{fig:running}, the inner loop's twelve program-counter cases of \texttt{ldmatrix} and \texttt{mma} become one body run four times, and the $K$-loop sixteen trips over it. $G_K$ calls the same $\mathrm{sem}_t$ as the flat model; only indexing and sequencing change, and a $\mathrm{for}$ can be treated by induction, one iteration at a time.

\textit{Verifying the segmentation.} The segmentation was written by an agent, so by our own rule it is not trusted; the stage ends with a Lean-checked obligation.
\begin{obligation}[Segmentation]\label{obl:seg}
$\mathrm{SEG}(K)$: for every $\sigma$ with $\mathrm{In}_I(\sigma, x)$, $F_K(\sigma) = G_K(\sigma)$.
\end{obligation}
This is an equality of partial functions, both faulting or both reaching the same state, proved phase by phase by induction on the folded indices from the lemmas the translator prepared (\Cref{sec:foundations}). If the segmentation is wrong, $\mathrm{SEG}(K)$ fails, at no cost in trust.

\textit{Proof loop.} An agent writes a proof, Lean checks it, and the errors go back to the agent; a proof passes only if it compiles without \texttt{sorry} and \texttt{\#print axioms} shows only Lean's standard axioms. When the attempts run out, a separate agent names the cause, a hard proof, a stuck agent, poorly chosen segments, or a wrong kernel, and the loop continues, restarts, redoes the segmentation, or returns the diagnosis to the kernel generator. This analysis only decides what to try next; whether a kernel is accepted, only Lean decides.

\textbf{Stage 3: Segment-wise verification loop.} The same loop now proves the segment-level model against the Torch semantics, the one place where the formalized module enters.
\begin{obligation}[Segment-wise equivalence]\label{obl:equiv}
$\mathrm{EQ}(K, S)$: for every $x$ and $\sigma$ with $\mathrm{In}_I(\sigma, x)$, there is a $\sigma'$ with $G_K(\sigma) = \sigma'$ and $\mathrm{Out}_I(\sigma') = \sem{S}(x)$.
\end{obligation}
Segment by segment, the agent shows what each phase and each $\mathrm{for}$ body leaves in the state, and induction carries it across the iterations. Together the obligations give \Cref{eq:equiv}, the theorem of \Cref{fig:running:thm}: $\mathrm{EQ}(K, S)$ supplies $\sigma'$ with $G_K(\sigma) = \sigma'$ and $\mathrm{Out}_I(\sigma') = \sem{S}(x)$, $\mathrm{SEG}(K)$ makes it $F_K(\sigma) = \sigma'$, and Stage 1 makes that $\sem{K}_M(\sigma) = \sigma'$.

\textbf{Statements.} The three stages prove a statement that no agent wrote, which is what the soundness claim rests on: a proof of the wrong statement proves nothing. Its premise and extractor are two definitions fixed once in $M$, applied to the task's \emph{signature} and launch:
\[
  \mathrm{In}_I := \mathrm{Pre}_M(\mathit{sig}, \mathit{launch}), \qquad \mathrm{Out}_I := \mathrm{Ext}_M(\mathit{sig}).
\]
$\mathrm{Pre}_M$ places every input buffer under its layout, keeps buffers apart and aligned, binds every scalar and dimension, and fixes the launch geometry; $\mathrm{Ext}_M$ reads every output buffer under its layout. The signature has one entry per kernel argument, a buffer with its role, element type, shape, and strided layout, or a scalar or dimension bound to the specification or pinned; it is the list the task's tests launch with, checked against the kernel's entry point (\Capp{app:schemas}). The theorem quantifies over the kernel's parameters (base addresses, strides, scalars), every state satisfying the premise, and every input, and the premise is proved satisfiable once for every signature, so no statement holds vacuously.

\subsection{What We Still Trust}\label{sec:trust}

Trust cannot be removed completely. What LARK does is narrow it, from the whole kernel to a short, fixed list, and report the list with every result. A certificate is
\[
  \mathcal{C} = (\Pi_M, \Pi_K, \vec\kappa, \pi), \qquad \Pi_K = (\mathit{sig}, \mathit{launch}),
\]
with the premises $\Pi = \Pi_M \cup \Pi_K$ of \Cref{sec:problem}, the fuels $\vec\kappa$ the fold runs under (data, not premises: exhausting one is a fault the proof rules out), and a Lean-checked proof
\begin{multline*}
  \pi \vdash\ \forall \vec p\, \sigma\, x.\ \mathrm{Pre}_M(\mathit{sig}, \mathit{launch})(\vec p, \sigma, x) \Rightarrow \mathrm{RF}(K)
  \ \wedge \\
  \exists \sigma'.\ \left(\mathrm{run}^{\vec\kappa}_{\mathrm{grid}}(\sigma) = \sigma' \wedge \mathrm{Ext}_M(\mathit{sig})(\vec p, \sigma') = \sem{S}(x)\right),
\end{multline*}
which with the reduction result of \Cref{sec:advanced} is $K \equiv_{M,I} S$. The two parts of $\Pi$ differ in what they are and how they are trusted.

$\Pi_M$ is kernel-independent, the statement that $M$ is faithful to the hardware $\mathcal{H}$ of \Cref{sec:problem}: the idealizations of $M$ (exact arithmetic as scope, the fold of \Cref{eq:fold} with the reduction result of \Cref{sec:advanced}), $\mathrm{sem}$ as measured against the hardware, the Lean formalization of the Torch module, and Lean itself (full list in \Capp{app:trusted-base}). 
The Torch formalization is the one agent-written artifact no proof covers: it is written independently of kernels, auditable on its own, and tested ahead of time against the module.

$\Pi_K$ is kernel-specific and small: the signature and the launch, which come from the task, not from the kernel or an agent (\Cref{sec:bridge}). What remains to believe is that they describe the task, which the report states by quoting them.

Floating-point behavior lies outside $\Pi$: no bound $|K_{\mathrm{fp}}(x) - \sem{S}(x)| \le \varepsilon$ is proved for the kernel $K_{\mathrm{fp}}$ as the hardware rounds it, and LARK claims nothing about rounding, overflow, or quantization; a GPU test against the Torch module answers that question, at a tolerance the report states beside the theorem.

Everything else is proved for the kernel at hand or guarded by refusal: $\mathrm{RF}(K)$; the absence of faults, convergence, and termination within the fuel, which together are $\sem{K}_M(\sigma){\downarrow}$; and $\mathrm{SEG}(K)$ and $\mathrm{EQ}(K, S)$. A theorem proved once in $M$ assembles these into $\pi$. Each run exports the certificate as a standalone Lean project that re-checks without LARK (\Capp{app:export}), with a \emph{trusted-assumption report}: $\Pi_M$ as a fixed list of statements, $\Pi_K$ quoted, the refusals that applied, and the sites where a named constant was read as exact. Trust is placed only where the report says it is.

\section{Experiments}
\label{sec:experiments}

\begin{table}
\centering
\caption{KernelCraft Level~1 results per target language and model: pass rate over the twelve tasks, and mean input and output tokens (millions) over the proven tasks, since a failed run spends to its budget. Per-task results in \Capp{app:l1-table}.}
\label{tab:main-results}
\setlength{\tabcolsep}{4pt}
\resizebox{\columnwidth}{!}{%
\begin{tabular}{@{}llcccc@{}}
\toprule
\makecell[l]{Target\\Lang.} & Metric & \makecell{GLM 5.3} & \makecell{GLM 5.3\\Flash} & \makecell{DeepSeek\\V4.1 Flash} & \makecell{GPT-6 Luna} \\
\midrule
\multirow{3}{*}{PTX}  & pass rate        & 83\%   & 58\%   & 25\%   & 25\% \\
                      & input tokens (M) & 314.53 & 714.60 & 148.47 & 445.93 \\
                      & output tokens (M)& 8.05   & 10.54  & 1.93   & 3.26 \\
\midrule
\multirow{3}{*}{CUDA} & pass rate        & 75\%   & 50\%   & 25\%   & 25\% \\
                      & input tokens (M) & 280.49 & 626.53 & 80.45  & 476.20 \\
                      & output tokens (M)& 6.18   & 9.09   & 1.12   & 3.42 \\
\midrule
\multirow{3}{*}{HIP}  & pass rate        & 83\%   & 50\%   & 25\%   & 25\% \\
                      & input tokens (M) & 312.21 & 773.96 & 156.15 & 289.17 \\
                      & output tokens (M)& 7.15   & 10.59  & 1.86   & 1.89 \\
\bottomrule
\end{tabular}%
}
\vspace{-1em}
\end{table}

\begin{table}
\centering
\caption{KernelCraft Level~2 results on PTX: a check mark is a certificate; tokens are input and output tokens (millions), in parentheses for runs that failed or stopped at their budget and left out of the Total average.}
\label{tab:verify-model}
\resizebox{\linewidth}{!}{%
\begin{tabular}{@{}llcccccc@{}}
\toprule
Model & Metric & FFN & SwiGLU & RoPE & FlashAttn & GQA & \textbf{Total} \\
\midrule
\multirow{3}{*}{Opus 5.5} & pass & \cmark & \cmark & \cmark & \cmark & \cmark & 100\% \\
 & input token (M) & 78.20 & 66.60 & 15.50 & 172.50 & 203.90 & 107.34 \\
 & output token (M) & 1.43 & 1.01 & 0.37 & 2.77 & 3.09 & 1.73 \\
\midrule
\multirow{3}{*}{GLM 5.3} & pass & \xmark & \xmark & \cmark & \xmark & \xmark & 20\% \\
 & input token (M) & (459.60) & (512.30) & 71.90 & (214.10) & (611.60) & 373.90 \\
 & output token (M) & (9.13) & (15.96) & 2.70 & (4.49) & (14.16) & 9.29 \\
\bottomrule
\end{tabular}%
}
\vspace{-1em}
\end{table} 

We show that LARK verifies most of the kernels a generation harness writes, each model stopping at a fixed rung of proof difficulty (\Cref{sec:eval-generated}); that a certificate is robust, rejecting single-site mutants (\Cref{sec:eval-mutation}); and that the method holds on production kernels, where it finds a real race (\Cref{sec:eval-library}). Ablations of the failure analysis and the segmentation (\Cref{sec:eval-discussion}) and the takeaways (\Cref{sec:eval-takeaways}) close the section.

\textbf{Setup.}
We evaluate LARK on KernelCraft~\citep{kernelcraft}, a recent benchmark of GPU kernel-generation tasks, each with a Torch reference module and tests, which gives LARK its specifications and the tests their baseline: the twelve Level~1 tasks, single operators, on PTX, CUDA, and HIP; and five Level~2 tasks, fused operators (FFN, SwiGLU, RoPE, FlashAttention, GQA), on PTX (\Cref{tab:kernelcraft-levels}).
To measure verification alone, one kernel is generated once and pinned for every task and language, and every model verifies the same kernel. 
The verification agents are four small models on Level~1 (GLM~5.3, GLM~5.3~Flash, DeepSeek~V4.1~Flash, GPT-6~Luna) and Opus~5.5 on Level~2, with GLM~5.3, the strongest of the four, as the comparison. 
Each verification runs under a cap on dialogue rounds per proof step and a spending ceiling, both read off preliminary runs (\Capp{app:setup}). \textbf{Metrics.} A verification passes when it ends in a certificate that Lean accepts with its standard axioms only, and the pass rate is the share of tasks that pass; cost is the input and output tokens of the verification.
The numeric screen and the latency measurements use an H100~80GB for PTX and CUDA and an MI300X for HIP.

\subsection{Verifying Generated Kernels: Difficulty Set by Proof Shape, Not Language}
\label{sec:eval-generated}

\emph{The tasks form a ladder of proof difficulty, and each model climbs to a fixed height.} \Cref{tab:main-results} gives the pass rates and token costs on Level~1 and \Cref{tab:verify-model} the results on Level~2 (per-task outcome details in \Capp{app:l1-table}). On all three languages every model proves the elementwise kernels; the two GLM models also prove the row-wise normalizations (GLM~5.3 stops at its spending ceiling on LayerNorm on HIP); only GLM~5.3 proves the tensor-core and convolution kernels; no model proves DepthwiseConv, and BatchMatmul is proved only on HIP, where the pinned kernel runs one 64-lane wavefront per block instead of four warps; and only Opus~5.5 proves the fused Level~2 kernels (pass rate and cost against model size in \Capp{app:model-size}). Each step up adds a kind of obligation, an induction over a loop, a convergence fact at a cut, a fragment-layout lemma, and a model that cannot close one kind fails every kernel that needs it. The two kernels that stop every model on PTX and CUDA stop them on difficulty alone: no statement is false, no verdict blames the kernel, no construct is refused, and every model stops on the same proof, one warp's tensor-core loop of 16 or 26 collective steps carried through the lane, warp, and block folds (\Capp{app:hard}). Proof cost follows the same shape: over the 116 PTX and CUDA runs it correlates with the length of the compiled program (Spearman 0.61), while the length barely predicts the number of agent turns (0.17), a turn being one proof attempt and its Lean check, so a long kernel costs more per turn, not more attempts (\Capp{app:scaling}). Every failure is a proof that ran out of rounds or budget, and on PTX and CUDA the two equivalence proofs take over 90\% of a run's tokens (\Capp{app:stage-cost}).

\emph{The target language matters less than the kernel, and the lower level is not the harder one.} Every model proves nearly the same set of kernels on PTX, CUDA, and HIP: the pass rates of a model differ by at most one task across the three languages, and the rung at which it stops is the same on each. The few cells that differ turn on the shape of the proof, not on the language. GEMV is the clearest case: the two GLM models prove it on PTX but not on CUDA, where the theorem is identical and every failure-analysis verdict was ``theorem true, proof hard'', but the kernel compiles to ten segments with two block barriers inside the $K$-loop against one segment with one barrier on PTX, so the PTX proof is shorter (29 agent turns against 48 for GLM~5.3). The shorter proof does not come from a weaker kernel: the verified PTX kernels are faster than the CUDA ones on half of the Level~1 kernels and within 5\% of them on the rest, and 1.8 times faster on GEMV (\Capp{app:latency}).

\subsection{Robustness: Rejecting Semantics-Breaking Mutants}
\label{sec:eval-mutation}

To test that a certificate is robust, we mutate verified kernels and ask whether LARK can correctly reject them. \Cref{tab:mutation-rules} lists the five mutation rules and the results; each mutant is a verified kernel with one change, on ReLU, Softmax, and GEMM in CUDA, PTX, and HIP. We test one semantics-breaking mutant per applicable rule and kernel; ReLU has no barrier, so R3 applies only to Softmax and GEMM. All runs use GLM~5.3 with the numeric screen disabled.
\begin{table}
\centering
\small
\caption{Mutation testing: per rule and target language, the mutants LARK rejects out of those tested. Every mutant is rejected; R3 has two mutants per language because ReLU has no barrier.}
\label{tab:mutation-rules}
\setlength{\tabcolsep}{3.5pt}
\renewcommand{\arraystretch}{1.1}
\resizebox{\columnwidth}{!}{%
\begin{tabular}{@{}c l >{\raggedright\arraybackslash}p{0.42\columnwidth} ccc@{}}
\toprule
& & & \multicolumn{3}{c@{}}{\textbf{Rejected / tested}} \\
\cmidrule(l){4-6}
\textbf{ID} & \textbf{Rule} & \textbf{Change} & \textbf{CUDA} & \textbf{PTX} & \textbf{HIP} \\
\midrule
R1 & Arithmetic & One operator or immediate ($+$ for $-$, a constant) & 3/3 & 3/3 & 3/3 \\
R2 & Load/store & One address, buffer, or stored value & 3/3 & 3/3 & 3/3 \\
R3 & Barrier & One barrier removed or moved one statement & 2/2 & 2/2 & 2/2 \\
R4 & Register & One variable or source register replaced at a use & 3/3 & 3/3 & 3/3 \\
R5 & Loop range & One loop bound, increment, or stride & 3/3 & 3/3 & 3/3 \\
\midrule
\multicolumn{3}{@{}l}{Total} & 14/14 & 14/14 & 14/14 \\
\bottomrule
\end{tabular}}
\vspace{-2em}
\end{table}

LARK rejects all 42 mutants: 11 at the translator's static checks, 10 by a refuted race-freedom obligation, 18 by the failure analysis showing that the kernel computes a different function from the specification, and 3 by a statement Lean refuses because the mutant needs a premise the task's signature does not give (\Capp{app:mutation_results}).

\subsection{Verifying Production Kernels}
\label{sec:eval-library}

\begin{table}
\centering
\caption{Verification of the ten production kernels (adapted as \Capp{app:library} records), each in its library's language (CUDA, except \texttt{vllm-awq-gemm} on PTX) by the model in its column. A check mark is a certificate; tokens are input and output tokens, in millions.}
\label{tab:lib-kernels}
\resizebox{\columnwidth}{!}{%
\begin{tabular}{@{}lccccc@{}}
\toprule
Metric & \multicolumn{1}{m{2cm}}{vllm-silu-and-mul} & \multicolumn{1}{m{2cm}}{vllm-rms-norm} & \multicolumn{1}{m{2cm}}{sglang-rmsnorm} & \multicolumn{1}{m{2cm}}{vllm-rotary-embedding} & \multicolumn{1}{m{2cm}}{llmc-softmax-forward} \\ \midrule
model used & GLM 5.3 & GLM 5.3 & GLM 5.3 & GLM 5.3 & GLM 5.3 \\
verified & \cmark & \cmark & \cmark & \cmark & \xmark \\
input tokens (M) & 192.90 & 294.00 & 529.00 & 152.80 & (0) \\
output tokens (M) & 4.56 & 8.12 & 12.23 & 4.44 & (0) \\ \midrule
Metric & \multicolumn{1}{m{2cm}}{sglang-fused-add-rmsnorm} & \multicolumn{1}{m{2cm}}{flashinfer-rmsnorm} & \multicolumn{1}{m{2cm}}{vllm-fused-qknorm-rope} & \multicolumn{1}{m{2cm}}{fam-flash-attn} & \multicolumn{1}{m{2cm}}{vllm-awq-gemm} \\ \midrule
model used & Opus 5.5 & Opus 5.5 & Opus 5.5 & Opus 5.5 & Opus 5.5 \\
verified & \cmark & \cmark & \cmark & \cmark & \cmark \\
input tokens (M) & 70.40 & 62.50 & 146.20 & 117.20 & 223.20 \\
output tokens (M) & 0.99 & 1.19 & 1.87 & 2.00 & 2.86 \\ \bottomrule
\end{tabular}%
}
\vspace{-2em}
\end{table}
We further evaluate LARK on ten production kernels adapted from vLLM~\citep{vllm}, SGLang~\citep{sglang}, FlashInfer~\citep{flashinfer}, llm.c~\citep{llmc}, and flash-attention-minimal~\citep{flashattnmin}: \emph{nine verify, and the tenth is rejected for a data race} (\Cref{tab:lib-kernels}). Each kernel enters the pipeline through small recorded rewrites, most of which reshape the code without changing what it computes; nine of the ten also carry a semantic \emph{departure}, such as a 64-bit index read as 32 bits, and each theorem is about the kernel with its departures, all listed in \Capp{app:library}. The five with the simplest shapes, the rejected one among them, ran with GLM~5.3, the five that fuse stages or run several phases with Opus~5.5.

The rejected kernel shows what a certificate catches that random-input tests do not. \texttt{llmc\_softmax\_forward} reads, in its last pass, a neighbouring output cell that another thread of the same block writes in the same pass, with no barrier between. The value read is discarded, so the race is benign and no test sees it. The translator cannot derive the footprint of the store, so the obligation falls to the agent, which finds a counterexample instead, two threads of one block storing and loading the same cell of \texttt{out} in the same phase, confirmed by the independent second opinion. The kernel is rejected, since the obligation is over footprints, not over whether a value read is used (\Capp{app:library}).

\subsection{Ablations: Failure Analysis and Segmentation}
\label{sec:eval-discussion}

Two parts of LARK are heuristics outside the trust boundary, and the runs show that both earn their place. The failure analysis, an analyst that names a cause after a failed round, a falsity probe, and an independent second analyst, only chooses what the loop tries next. \emph{It is cheap, and it never cut off a proof that was going to succeed.} Over the Level~1 runs on PTX and CUDA, the Level~2 runs, and the production kernels (\Capp{app:analyst}), the analysis costs 0.7\% to 4.7\% of a model's input tokens and ended ten dialogues before their ceiling, and in the preliminary runs, where it ran without acting, no proof closed after the point where it would have stopped. It is fallible: it called a correct kernel buggy three times, each refuted by the second analyst; in all, the second analyst disputed eleven of the twenty stopping verdicts it reviewed. None of this touches soundness, since a wrong verdict can only end a run early.

\emph{On a complicated kernel the segmentation saves more than its own equivalence proof costs.} We conducted an ablation that proves the specification directly against the flat model, with no segmentation and no first equivalence, on FlashAttention on PTX with Opus~5.5. The direct proof closed at 220M input tokens over 20 turns, against 120M over 11 for the second equivalence with the segmentation; with the 49M tokens of the segmentation and its equivalence counted, the direct run still used 29\% more and crossed a ceiling the default run finished under, because without segments the agent re-proves the same loop body at every program counter (\Capp{app:flat}).

\subsection{Discussion}\label{sec:eval-takeaways}

The experiments point to three properties of the framework.

\textbf{1) Automated formal verification keeps pace with generation.} GLM~5.3 certifies 10 of the 12 Level-1 kernels on PTX and on HIP (9 on CUDA) and Opus~5.5 every Level-2 kernel with no human in the loop, which shows that agentic formal verification of GPU kernels is feasible, and the cost per proof falls as the models improve.

\textbf{2) A robust pipeline with soundness guaranteed.} Lean supplies the soundness: a certificate exists only when every obligation is proved, so no agent can talk its way to one. LARK is also robust in practice: wrong segmentations and wrong analyst verdicts ended as failed proofs, semantics-breaking mutants are rejected (\Cref{sec:eval-mutation}), and the one production kernel with a race was rejected while the nine correct ones were certified (\Cref{sec:eval-library}).

\textbf{3) A narrowed trust boundary.} What a certificate rests on is independent of the kernel: the machine model, the operation semantics measured against the hardware, and the premise and extractor definitions are fixed and checked offline, before any kernel is seen, the specification is formalized once per task and tested, and everything about the kernel itself is proved.
LARK is thus a step toward verifiable code generation in the agentic era: the generator stays untrusted, agents write the proofs, and the certificates are sound.

\section{Related Work}
\label{sec:related}
Kernel generators and benchmarks~\citep{kernelbench, kevin, kernelcraft, cake} judge a kernel correct when its outputs match a reference on test inputs, a check that generated kernels have been shown to exploit~\citep{lange2025robustagenticcudakernel, woo2025tritonrl}. Stricter tests~\citep{correctnessillusion, contractgradeverifier} catch more errors, but on every untested input the kernel's correctness is still assumed. Earlier formal tools either prove only race freedom~\citep{gpuverify} or rely on human proofs for each program~\citep{cudaaucoq, lnsym}.

Two recent verifiers have been applied to LLM-generated kernels. Volta~\citep{volta} decides equivalence between a PTX kernel and a reference PTX kernel over the reals. Its specification is thus itself a low-level kernel that must be correct, its decision procedure is not mechanically verified, and its evaluation checks only the first pair of thread blocks, which covers the full grid only if every block behaves alike. ProofWright~\citep{proofwright} has LLM agents annotate CUDA kernels for the VerCors verifier~\citep{armborst2024vercors}, which proves memory and thread safety, but proves equivalence only for element-wise kernels with one output per thread, relying on manual preprocessing of kernels and hand-checked LLM annotations.

In both systems, what must be trusted includes assumptions about, or manual work on, the kernel itself. LARK's trust boundary is instead fixed before any kernel is seen: for each kernel, it proves in Lean equivalence to the task's Torch module under exact arithmetic, together with race freedom, absence of faults, and termination, or rejects the kernel.
\Capp{app:related} discusses further related work.

\section{Conclusion}
\label{sec:conclusion}

We propose LARK, a framework that verifies LLM-generated GPU kernels against their Torch specifications in Lean, with the trust boundary moved off the kernel. It gives generated code the check it lacked: one that keeps pace with generation and means something when it passes. The machine model and the operation semantics are fixed and checked once per target language; everything about the kernel, race freedom above all, is proved per kernel or the kernel is rejected; and agents may write the segmentations and the proofs because Lean checks every line, so a wrong agent output is a failed proof, never a wrong theorem. GLM~5.3 certifies 10 of the 12 Level-1 kernels and Opus~5.5 all five fused Level-2 kernels, and nine of ten adapted production kernels verify. The tenth, a softmax kernel from llm.c, is rejected for a real data race present in the shipped code, a bug invisible to input--output testing because the racing value is discarded, which LARK located from the code alone.

\subsection*{Limitations} 
\label{sec:limitations}

LARK's guarantee is bounded in three ways. Arithmetic is idealized: floats and integers are exact, $+\infty$ and NaN are faults, and the theorem says nothing about rounding, overflow, accumulation order, or the precision of approximate instructions; floating-point behavior is tested, not proved (\Cref{sec:trust}). Concurrency is phase-structured: blocks that communicate within a phase, benign races, and warps that halt early are rejected. The operation set is the pinned tables: asynchronous operations (TMA, \texttt{wgmma}, \texttt{mbarrier}), atomics, partial masks, bf16, 8-bit types, non-polynomial addresses, and multi-dimensional blocks are refused (\Capp{app:unsupported}). None of these changes the method; each costs a pinned definition or a lemma emitter, and retargeting to a new language is manual (\Capp{app:retarget}).





\bibliography{references}
\bibliographystyle{mlsys2025}


\fi  

\ifdefined\appendixonlyflag\def\includeappendix{}\fi
\ifdefined\preprintflag\def\includeappendix{}\fi
\ifdefined\includeappendix
\ifdefined\appendixonlyflag
\renewcommand{\thefigure}{S\arabic{figure}}   
\renewcommand{\thetable}{S\arabic{table}}
\renewcommand{\theequation}{S\arabic{equation}}
\fi
\appendix

\section{Appendix}

\subsection{Related Works}

\paragraph{Kernel generation and test-based checking.}
Existing LLM kernel generation across target languages, from CUDA to close-to-metal assembly~\citep{kernelbench, kevin, kernelcraft, cake}, checks correctness by comparing a kernel's outputs with a reference implementation on sampled inputs. Such testing has proved unreliable: kernels that pass KernelBench's checks have achieved fake speedups of $50$--$120\times$ by exploiting the evaluation harness, and excluding the affected tasks reduces the reported aggregate speedup from $3.13\times$ to $1.49\times$~\citep{lange2025robustagenticcudakernel}. Reward hacking is likewise a central obstacle for generators trained with reinforcement learning~\citep{woo2025tritonrl}. Stricter test suites, contract-based checks, and metamorphic testing~\citep{correctnessillusion, contractgradeverifier, metamorphictesting} catch more errors, but their results remain empirical: a kernel that passes them has been checked only on the inputs that were tried, and its correctness on every other input is assumed. The kernel therefore remains inside the trust boundary.

\paragraph{Formal verification and equivalence checking.}
Translation validation~\citep{translationvalidation, translationval4oc, alive2}, equivalence checking of program transformations~\citep{peqc-fpga24}, and verified compilers~\citep{compcert} relate a program to the source from which it was derived. These techniques do not apply directly to LLM-generated kernels, which have no such source and are specified only by a high-level module in a different language. Interactive proofs against a formal ISA semantics~\citep{cudaaucoq, lnsym} exclude the kernel from the trust boundary, but require substantial manual effort for each program. Conversely, GPUVerify~\citep{gpuverify} is fully automatic but establishes only the absence of data races and barrier divergence; since it does not relate a kernel to its specification, the kernel's functional behavior remains within the trust boundary. LARK combines both properties: it is automatic, treats race freedom as one proof obligation among several, and additionally proves that the kernel is equivalent to its specification.

\paragraph{LLM-driven proving and its failure modes.}
LLM agents can now write machine-checked proofs~\citep{multimodelverifier}. A checked proof, however, is only as trustworthy as the statement it proves, and statements are where errors concentrate: audits of Lean benchmarks find missing hypotheses, simplified problems, and vacuous theorems~\citep{faultsformalbench}, and models asked to annotate a program for verification have been observed to modify the program instead~\citep{banerjee2026dafnypro}. When the agent that writes the proof also writes the statement, the statement lies inside the trust boundary, and so does the agent. LARK keeps agents out of the per-kernel trust boundary: the theorem is fixed by the task before the kernel is seen, and Lean checks every proof an agent writes, so an agent's error about the kernel can only cause a proof to fail, never a certificate for the wrong theorem. The one agent-written artifact that remains trusted, the specification formalization, is per task, tested, and reported as such (\Cref{sec:trust}).

\paragraph{Formal verification of LLM-generated GPU kernels.}
Two recent verifiers have been applied to LLM-generated kernels. Volta~\citep{volta}, a general equivalence checker, uses symbolic execution to decide equivalence between a PTX kernel and a reference PTX kernel over the reals. Its specification is thus itself a low-level kernel that must be correct, its decision procedure is not mechanically verified, and its evaluation checks only the first pair of thread blocks, which covers the full grid only if every block behaves alike. ProofWright~\citep{proofwright} has LLM agents annotate CUDA kernels for the deductive verifier VerCors~\citep{armborst2024vercors}, which proves memory and thread safety, but proves equivalence only for element-wise kernels with one output per thread, relying on manual preprocessing of each kernel and hand-checked LLM annotations. LARK instead checks the kernel against the task's Torch module, fixes its trust boundary before any kernel is seen, and proves in Lean equivalence under exact arithmetic, together with race freedom, absence of faults, and termination, or rejects the kernel.
\label{app:related}

\subsection{Experiment Setup Details}
\label{app:setup}

\Cref{tab:kernelcraft-levels} lists the evaluated tasks by KernelCraft level, \Cref{tab:kernelcraft-model-budgets} the round caps and spending ceilings per model, and \Cref{tab:kernel-levels} the kernels by the highest cut level of their phase structure.
\newcolumntype{L}[1]{>{\raggedright\arraybackslash}p{#1}}

\begin{table}
\centering
\small
\caption{The evaluated KernelCraft tasks by level.}
\label{tab:kernelcraft-levels}
\setlength{\tabcolsep}{8pt}
\renewcommand{\arraystretch}{1.25}
\begin{tabular}{@{} L{0.26\columnwidth} L{0.64\columnwidth} @{}}
\toprule
\textbf{KernelCraft Level} & \textbf{Tasks} \\
\midrule
Level 1: Primitive Operations &
\texttt{silu}, \texttt{relu}, \texttt{gelu},
\texttt{softmax}, \texttt{layernorm}, \texttt{rmsnorm},
\texttt{gemv}, \texttt{gemm}, \texttt{batchmatmul},
\texttt{linear}, \texttt{conv2d}, \texttt{depthwise\_conv} \\
\addlinespace
Level 2: Composite Operations &
\texttt{rope}, \texttt{ffn}, \texttt{swiglu}, \texttt{flash\_attn}, \texttt{gqa} \\
\bottomrule
\end{tabular}
\end{table}

\begin{table*}
\centering
\small
\caption{Per-verification budgets by model and task group: the cap on dialogue rounds per proof step (\texttt{-{}-max-auto-tries}) and the spending ceiling (\texttt{-{}-max-usd}). The round caps were read off preliminary runs in which the failure analyst always answered CONTINUE (GEMM on PTX with DeepSeek V4.1 Flash and GLM 5.3, GQA on PTX with Opus 5.5): the cap is the round of the last proof closure at the 75th to 90th percentile, plus one. The ceiling is the dollar amount at which a run was stopped; it is given for reference only.}
\label{tab:kernelcraft-model-budgets}
\setlength{\tabcolsep}{6pt}
\renewcommand{\arraystretch}{1.2}
\begin{tabularx}{\textwidth}{@{} L{0.14\textwidth} >{\raggedright\arraybackslash}X ccccc @{}}
\toprule
& & \multicolumn{5}{c@{}}{\textbf{Rounds / ceiling (\$)}} \\
\cmidrule(l){3-7}
\textbf{Group} & \textbf{Workloads} &
\makecell{DeepSeek\\V4.1 Flash} & \makecell{GPT-6\\Luna} & \makecell{GLM 5.3\\Flash} & \makecell{GLM 5.3} & \makecell{Opus 5.5} \\
\midrule
L1
& \texttt{relu}, \texttt{silu}, \texttt{gelu}, \texttt{softmax}, \texttt{layernorm}, \texttt{rmsnorm}, \texttt{gemv}, \texttt{gemm}, \texttt{linear}, \texttt{batchmatmul}, \texttt{conv2d}, \texttt{depthwise\_conv}
& 16 / 50 & 16 / 50 & 16 / 75 & 8 / 100 & --- \\
\addlinespace
L2
& \texttt{rope}
& --- & --- & --- & 16 / 100 & 10 / 100 \\ \addlinespace

& \texttt{ffn}, \texttt{swiglu}, \texttt{flash\_attn}, \texttt{gqa}
& --- & --- & --- & 16 / 100 & 10 / 200 \\
\addlinespace
Library
& \texttt{vllm\_silu\_and\_mul}, \texttt{sglang\_rmsnorm}, \texttt{vllm\_rms\_norm}, \texttt{vllm\_rotary\_embedding}
& --- & --- & --- & 8 / 100 & --- \\ \addlinespace

& \texttt{llmc\_softmax\_forward}, \texttt{sglang\_fused\_add\_rmsnorm}, \texttt{flashinfer\_rmsnorm}, \texttt{vllm\_fused\_qknorm\_rope}
& --- & --- & --- & --- & 10 / 100 \\ \addlinespace

& \texttt{fam\_flash\_attn}, \texttt{vllm\_awq\_gemm}
& --- & --- & --- & --- & 10 / 200 \\
\bottomrule
\end{tabularx}
\end{table*}

\begin{table}
\centering
\small
\caption{The evaluation kernels by the highest cut level of their phase structure (\Cmain{sec:foundations}): the 17 kernels of the core set (KernelCraft Levels 1 and 2) and the 10 production kernels. A kernel with no cut runs as one phase; a warp-collective operation alone does not add a level.}
\label{tab:kernel-levels}
\setlength{\tabcolsep}{5pt}
\renewcommand{\arraystretch}{1.2}
\begin{tabular}{@{} >{\raggedright\arraybackslash}p{0.17\columnwidth} >{\raggedright\arraybackslash}p{0.38\columnwidth} >{\raggedright\arraybackslash}p{0.36\columnwidth} @{}}
\toprule
\textbf{Highest cut} & \textbf{Core set} & \textbf{Library kernels} \\
\midrule
None &
\texttt{relu}, \texttt{silu}, \texttt{gelu}, \texttt{conv2d}, \texttt{rope} &
\texttt{vllm\_silu\_\allowbreak and\_\allowbreak mul}, \texttt{vllm\_rotary\_\allowbreak embedding} \\
\addlinespace
Warp &
--- &
\texttt{vllm\_fused\_\allowbreak qknorm\_\allowbreak rope} \\
\addlinespace
Block &
\texttt{softmax}, \texttt{layernorm}, \texttt{rmsnorm}, \texttt{gemv}, \texttt{gemm}, \texttt{linear}, \texttt{batchmatmul}, \texttt{depthwise\_conv}, \texttt{flash\_attn}, \texttt{gqa} &
\texttt{sglang\_rmsnorm}, \texttt{vllm\_rms\_norm}, \texttt{flashinfer\_\allowbreak rmsnorm}, \texttt{sglang\_fused\_\allowbreak add\_\allowbreak rmsnorm}, \texttt{llmc\_softmax\_\allowbreak forward}, \texttt{fam\_flash\_attn}, \texttt{vllm\_awq\_gemm} \\
\addlinespace
Grid &
\texttt{ffn}, \texttt{swiglu} &
--- \\
\bottomrule
\end{tabular}
\end{table}

\paragraph{Preliminary runs.} The round caps of \Cref{tab:kernelcraft-model-budgets} were read off preliminary verify-only runs in which every failed round continued the same dialogue, under a cap of 15 rounds, while the failure analyst still ran and recorded what the normal gate would have done. A round \emph{closes an obligation} when the agent's proof of that round discharges one of the open lemmas or goals of the equivalence, and the rule is a high percentile of the last closing round, plus one round of margin. DeepSeek~V4.1~Flash ran GEMM on PTX eight times, none proving the first equivalence, with last closing rounds at 1, 5, 8, 10, 11, 12, 23, and 23; the 75th percentile gives a cap of 16, which GPT-6~Luna and GLM~5.3~Flash, without preliminary runs of their own, take by decision. GLM~5.3 ran GEMM on PTX three times and proved both equivalences each time, with last closing rounds 1, 1, and 3 for the first and 5, 7, and 6 for the second, a cap of 8. Opus~5.5 ran GQA on PTX once and closed the second equivalence in round 9, a cap of 10. The spending ceilings are the amounts at which these runs were stopped. None of these runs is an evaluation cell.

\Cref{tab:task-configs} gives the configuration of each evaluated task in the notation of KernelCraft; every task runs at the largest of the benchmark's five configurations.

\begin{table}
\centering
\small
\caption{Configuration of each evaluated task, in the notation of KernelCraft~\citep{kernelcraft}: every task runs at the largest of the benchmark's five configurations, and all tensors are FP16. $n$ = number of elements (input tokens), $M, K, N$ = matrix dimensions, $b$ = batch size, $h_{\mathrm{in}}, h_{\mathrm{out}}$ = input and output features, $h$ = hidden size, $h_i$ = intermediate size, $s$ = sequence length, $n_h$ = attention heads, $n_{kv}$ = key-value heads, $d_{\mathrm{head}}$ = head dimension, $H$ = spatial size (height $=$ width), $c_{\mathrm{in}}, c_{\mathrm{out}}$ = input and output channels, $c$ = channels, $k$ = kernel size.}
\label{tab:task-configs}
\setlength{\tabcolsep}{6pt}
\renewcommand{\arraystretch}{1.15}
\begin{tabular}{@{} l l @{}}
\toprule
\textbf{Task (parameters)} & \textbf{Configuration} \\
\midrule
\multicolumn{2}{@{}l}{\emph{Level 1}} \\
SiLU, ReLU, GELU $(n)$ & $(16384)$ \\
Softmax, LayerNorm, RMSNorm $(n)$ & $(16384)$ \\
GEMV $(M, N)$ & $(512, 1024)$ \\
GEMM $(M, K, N)$ & $(64, 256, 512)$ \\
BatchMatmul $(b, M, K, N)$ & $(32, 64, 256, 512)$ \\
Linear $(n, h_{\mathrm{in}}, h_{\mathrm{out}})$ & $(64, 256, 512)$ \\
Conv2D $(H, c_{\mathrm{in}}, c_{\mathrm{out}}, k)$ & $(8, 32, 8, 3)$ \\
DepthwiseConv $(H, c, k)$ & $(16, 32, 5)$ \\
\addlinespace
\multicolumn{2}{@{}l}{\emph{Level 2}} \\
FFN, SwiGLU $(n, h, h_i)$ & $(64, 256, 1024)$ \\
RoPE $(s, d_{\mathrm{head}})$ & $(256, 128)$ \\
FlashAttention $(s, h)$ & $(128, 256)$ \\
GQA $(s, h, n_h, n_{kv})$ & $(128, 512, 8, 2)$ \\
\bottomrule
\end{tabular}
\end{table}

\subsection{Experiment Configuration}
\newcolumntype{L}[1]{>{\raggedright\arraybackslash}p{#1}}
\newcolumntype{Y}{>{\raggedright\arraybackslash}X}
\newcolumntype{S}[1]{>{\leavevmode\footnotesize\color{black!70}\raggedright\arraybackslash}p{#1}}

\begin{table*}
\centering
\footnotesize
\caption{Configuration of the verification runs, per model. Every run passes \texttt{-{}-verify-only}, so the kernel is the registry's pinned program for its task and language and the generation flags are unused. ``default'' means the flag is not passed. Round caps and spending ceilings are per verification of one kernel; the ceilings are in dollars because they are what the runs were stopped at, and their derivation from the preliminary runs is in \Cref{tab:kernelcraft-model-budgets}. The last column says what each flag controls; the settings shared by every run are listed below the per-model rows.}
\label{tab:final-experiment-parameters}
\setlength{\tabcolsep}{4pt}
\renewcommand{\arraystretch}{1.1}
\begin{tabularx}{\textwidth}{@{} >{\raggedright\arraybackslash}p{0.13\textwidth} Y Y Y Y Y >{\footnotesize\color{black!70}\raggedright\arraybackslash}p{0.2\textwidth} @{}}
\toprule
\textbf{Flag} & \textbf{DeepSeek V4.1 Flash} & \textbf{GPT-6 Luna} & \textbf{GLM 5.3 Flash} & \textbf{GLM 5.3} & \textbf{Opus 5.5} & \textbf{\color{black}What it controls} \\
\midrule
\texttt{-{}-model} & \texttt{deepseek-\allowbreak v4.1-\allowbreak flash-\allowbreak1017} & \texttt{gpt-6-luna} & \texttt{glm-5.3-\allowbreak flash} & \texttt{glm-5.3} & \texttt{opus-5.5} & Model of the verification agents \\
\texttt{-{}-isa} & PTX, CUDA, HIP & PTX, CUDA, HIP & PTX, CUDA, HIP & L1: all three; L2: PTX; library: CUDA & L2: PTX; library: CUDA, PTX & Target language of the run \\
\texttt{-{}-kernel} & L1 & L1 & L1 & L1, L2, production libraries & L2, production libraries & Task verified (\Cref{tab:kernelcraft-levels}) \\
\addlinespace
\texttt{-{}-backward-\allowbreak effort} & \texttt{xhigh} & \texttt{xhigh} & default & default & \texttt{xhigh} & Reasoning effort of every verification agent \\
\texttt{-{}-proof-\allowbreak effort} & \texttt{xhigh} & \texttt{xhigh} & default (\texttt{max}) & default (\texttt{max}) & \texttt{xhigh} & Effort of the two proof agents \\
\texttt{-{}-other-\allowbreak effort} & default (\texttt{xhigh}) & default (\texttt{xhigh}) & default (\texttt{high}) & default (\texttt{high}) & default (\texttt{xhigh}) & Effort of the statement, review, and diagnosis agents \\
\addlinespace
\texttt{-{}-max-auto-\allowbreak tries} & 16 & 16 & 16 & 8 (L1, library); 16 (L2) & 10 & Round cap per proof \\
\texttt{-{}-max-usd} & 50 & 50 & 75 & 100 & 100 or 200 by task & Spending ceiling per verification \\
\midrule
\multicolumn{7}{@{}>{\raggedright\arraybackslash}p{\textwidth}@{}}{\textbf{Shared by every run.}
\texttt{-{}-ask-on-max-usd} on (stop at the ceiling rather than continue);
\texttt{-{}-repeats} 2 (pipeline restarts allowed);
\texttt{-{}-translation-attempts} 10 (attempts per segmentation or statement step);
\texttt{-{}-proof-attempts} 10 (attempts per proof step);
\texttt{-{}-check-attempts} 3 (attempts per review);
\texttt{-{}-diagnosis-attempts} 3 (attempts per failure analysis);
\texttt{-{}-lean-timeout} 150\,s (time limit per Lean check);
\texttt{-{}-max-lean-tool-calls} 40 (Lean checks per agent turn);
\texttt{-{}-numeric-trials} 6 and \texttt{-{}-numeric-probes} 4 (random inputs of the numeric screen);
all static translator checks on;
\texttt{-{}-racefree-fallback} on (the agent proves race freedom where no lemma is generated);
\texttt{-{}-flat-model} off (direct proof without segmentation, ablation only).} \\
\bottomrule
\end{tabularx}
\end{table*}

Experiment configurations are detailed in \Cref{tab:final-experiment-parameters}.

\subsection{Additional Results}
\label{app:results}
Unless a subsection says otherwise, the analyses below are over the PTX and CUDA runs; the HIP runs enter only the per-task table and the latency measurements.

\subsubsection{Per-Task Results on Level 1}\label{app:l1-table} \Cref{tab:eval-main-res} gives, for every model, language, and task, the outcome and the input and output tokens of the verification. A run ends ``pass'' with a certificate, ``budget'' at its spending ceiling, and ``fail'' when the agent gave up or the round cap was hit.
\begin{table*}
\centering
\caption{Verification of the twelve KernelCraft GPU Level~1 tasks, per model and target language. A run ends ``pass'' with a certificate, ``budget'' at its spending ceiling, or ``fail'' at its round cap or by an agent failure. Tokens are input and output tokens of the verification, in millions; the ``Total'' column gives the pass rate and the average tokens over the proven kernels. A count in parentheses belongs to a run that did not prove its kernel, reflects the budget it was given rather than the cost of a proof, and is left out of the average.}
\label{tab:eval-main-res}
\resizebox{\linewidth}{!}{%
\begin{tabular}{@{}lllccccccccccccc@{}}
\toprule
Model & Target Lang. & Metric & ReLU & SiLU & GELU & Softmax & LayerNorm & RMSNorm & GEMV & GEMM & Linear & BatchMatmul & Conv2D & DepthwiseConv & \textbf{Total} \\ \midrule
\multirow{9}{2cm}{GLM 5.3} & \multirow{3}{*}{CUDA} & pass/fail/budget & pass & pass & pass & pass & pass & pass & budget & pass & pass & budget & pass & budget & 75\% \\
 &  & input tokens (M) & 44.67 & 61.61 & 87.60 & 332.71 & 291.45 & 481.23 & (700.51) & 389.51 & 430.82 & (549.06) & 404.79 & (562.16) & 280.49 \\
 &  & output tokens (M) & 1.78 & 1.78 & 2.38 & 7.97 & 6.32 & 5.93 & (15.60) & 8.95 & 11.25 & (13.53) & 9.23 & (13.47) & 6.18 \\
 & \multirow{3}{*}{PTX} & pass/fail/budget & pass & pass & pass & pass & pass & pass & pass & pass & pass & budget & pass & budget & 83\% \\
 &  & input tokens (M) & 119.20 & 77.30 & 56.50 & 391.20 & 483.20 & 288.60 & 481.00 & 277.70 & 508.70 & (752.70) & 461.90 & (616.80) & 314.53 \\
 &  & output tokens (M) & 2.66 & 2.77 & 1.91 & 10.63 & 11.50 & 6.68 & 13.79 & 8.09 & 10.54 & (15.33) & 11.94 & (15.93) & 8.05 \\
 & \multirow{3}{*}{HIP} & pass/fail/budget & pass & pass & pass & pass & budget & pass & pass & pass & pass & pass & pass & budget & 83\% \\
 &  & input tokens (M) & 158.84 & 83.79 & 99.16 & 229.01 & (692.63) & 231.08 & 576.31 & 422.98 & 414.93 & 455.50 & 450.52 & (581.50) & 312.21 \\
 &  & output tokens (M) & 4.33 & 2.53 & 4.62 & 6.45 & (17.44) & 7.93 & 12.31 & 11.22 & 9.54 & 9.93 & 2.66 & (15.53) & 7.15 \\
\midrule
\multirow{9}{2cm}{GLM 5.3 Flash} & \multirow{3}{*}{CUDA} & pass/fail/budget & pass & pass & pass & pass & pass & pass & budget & budget & budget & budget & budget & fail & 50\% \\
 &  & input tokens (M) & 221.00 & 281.60 & 248.70 & 669.40 & 457.40 & 1881.10 & (1981.50) & (2052.50) & (1901.70) & (2036.30) & (1811.90) & (1366.40) & 626.53 \\
 &  & output tokens (M) & 3.13 & 5.10 & 4.74 & 10.09 & 6.99 & 24.48 & (24.19) & (25.26) & (23.66) & (27.72) & (23.01) & (20.11) & 9.09 \\
 & \multirow{3}{*}{PTX} & pass/fail/budget & pass & pass & pass & pass & pass & pass & pass & fail & fail & budget & fail & budget & 58\% \\
 &  & input tokens (M) & 184.70 & 260.80 & 231.10 & 792.90 & 1455.90 & 1219.50 & 857.30 & (1408.60) & (1486.40) & (1974.40) & (1531.10) & (1891.60) & 714.60 \\
 &  & output tokens (M) & 3.14 & 4.05 & 3.51 & 11.91 & 20.94 & 16.38 & 13.85 & (15.64) & (21.43) & (30.21) & (25.58) & (24.16) & 10.54 \\
 & \multirow{3}{*}{HIP} & pass/fail/budget & pass & pass & pass & pass & pass & pass & budget & budget & budget & budget & budget & budget & 50\% \\
 &  & input tokens (M) & 240.88 & 271.10 & 292.31 & 841.58 & 990.15 & 2007.71 & (1019.44) & (1638.08) & (1880.69) & (1783.35) & (1477.32) & (1871.22) & 773.96 \\
 &  & output tokens (M) & 3.41 & 4.55 & 4.08 & 12.04 & 14.31 & 25.13 & (15.53) & (15.42) & (20.43) & (29.31) & (23.78) & (19.54) & 10.59 \\
\midrule
\multirow{9}{2cm}{DeepSeek V4.1 Flash} & \multirow{3}{*}{CUDA} & pass/fail/budget & pass & pass & pass & fail & fail & fail & fail & fail & fail & fail & fail & fail & 25\% \\
 &  & input tokens (M) & 95.39 & 66.84 & 79.12 & (458.90) & (780.60) & (711.59) & (528.54) & (706.41) & (744.72) & (550.34) & (403.48) & (574.90) & 80.45 \\
 &  & output tokens (M) & 1.19 & 0.96 & 1.20 & (7.63) & (10.80) & (9.62) & (6.92) & (10.27) & (11.85) & (8.65) & (6.11) & (8.16) & 1.12 \\
 & \multirow{3}{*}{PTX} & pass/fail/budget & pass & pass & pass & fail & fail & fail & fail & fail & fail & fail & fail & fail & 25\% \\
 &  & input tokens (M) & 224.58 & 101.24 & 119.58 & (368.11) & (197.33) & (383.49) & (367.04) & (653.25) & (485.78) & (215.81) & (456.24) & (430.42) & 148.47 \\
 &  & output tokens (M) & 2.60 & 1.54 & 1.64 & (4.46) & (3.50) & (5.06) & (5.46) & (9.09) & (7.30) & (3.05) & (6.85) & (6.21) & 1.93 \\
 & \multirow{3}{*}{HIP} & pass/fail/budget & pass & pass & pass & fail & fail & fail & fail & fail & fail & fail & fail & fail & 25\% \\
 &  & input tokens (M) & 241.67 & 133.71 & 93.07 & (471.13) & (831.19) & (741.90) & (462.83) & (688.31) & (691.33) & (390.18) & (501.33) & (812.59) & 156.15 \\
 &  & output tokens (M) & 1.98 & 2.01 & 1.59 & (6.32) & (10.91) & (8.91) & (5.53) & (10.10) & (8.82) & (2.98) & (7.03) & (9.03) & 1.86 \\
\midrule
\multirow{9}{2cm}{GPT-6 Luna} & \multirow{3}{*}{CUDA} & pass/fail/budget & pass & pass & pass & fail & fail & fail & fail & fail & fail & fail & fail & fail & 25\% \\
 &  & input tokens (M) & 690.80 & 316.40 & 421.40 & (638.30) & (707.80) & (1243.20) & (511.80) & (1171.60) & (596.90) & (660.10) & (689.10) & (1152.50) & 476.20 \\
 &  & output tokens (M) & 4.91 & 2.40 & 2.95 & (4.38) & (5.30) & (8.50) & (4.27) & (9.01) & (4.17) & (4.90) & (5.12) & (7.89) & 3.42 \\
 & \multirow{3}{*}{PTX} & pass/fail/budget & pass & pass & pass & fail & fail & fail & fail & fail & fail & fail & fail & fail & 25\% \\
 &  & input tokens (M) & 366.30 & 401.70 & 569.80 & (1144.90) & (1281.40) & (1494.60) & (547.50) & (1259.00) & (1146.20) & (1052.00) & (932.00) & (880.90) & 445.93 \\
 &  & output tokens (M) & 2.30 & 3.16 & 4.31 & (7.77) & (8.54) & (9.66) & (4.08) & (8.60) & (7.94) & (7.48) & (7.30) & (5.78) & 3.26 \\
 & \multirow{3}{*}{HIP} & pass/fail/budget & pass & pass & pass & fail & fail & fail & fail & fail & fail & fail & fail & fail & 25\% \\
 &  & input tokens (M) & 345.00 & 286.00 & 236.50 & (780.39) & (798.14) & (1339.64) & (580.62) & (1150.13) & (513.32) & (710.28) & (842.19) & (1451.10) & 289.17 \\
 &  & output tokens (M) & 2.13 & 2.06 & 1.48 & (4.71) & (7.71) & (9.05) & (4.31) & (8.33) & (6.95) & (6.93) & (6.20) & (8.68) & 1.89 \\ \bottomrule 
\end{tabular}%
}
\end{table*}

\subsubsection{Refusal Census}\label{app:refusals} Over the 116 PTX and CUDA verifications of the three subsets, no correct kernel was refused by scope: every kernel that passed the numeric screen compiled into the model and reached the proof loop. Outcomes on Level~1, per model (proved / budget stop / agent failure out of 24): GLM~5.3 19 / 5 / 0; GLM~5.3~Flash 13 / 7 / 4; GPT-6~Luna 6 / 0 / 18; DeepSeek~V4.1~Flash 6 / 0 / 18. On Level~2 (PTX): Opus~5.5 5 / 0 / 0; GLM~5.3 1 / 3 / 1. Library: 10 / 0 / 0.

\subsubsection{Cost per Stage}\label{app:stage-cost} Each verification splits into a static part (the translator's compile and its Lean modules, the certificate), the agent steps (segmentation, statements, the two proofs, reviews, and the diagnosis), and Lean checks inside the proof loop. \Cref{tab:stage-cost} gives the tokens per stage. On Level~1 the two equivalence proofs take 95 to 99\% of a run's tokens (median per model), the diagnosis 0.7 to 4.4\%, and the segmentation, statement, and review steps at most 0.3\% each; the segmentation proof dominates for the three weaker models, which rarely reach the equivalence proof, and the equivalence proof dominates for GLM~5.3 and on Level~2. The static part takes 36 to 157 seconds per run (maximum 352); Lean inside the loop 18 to 60 minutes per run for the small models and 6 to 8 for Opus~5.5.
\begin{table*}
\centering
\footnotesize
\caption{Tokens per stage of a verification (millions), per model and evaluation set: the median over the runs that reached the stage and, in parentheses, the median share of the run's tokens; ``$k$ of $N$'' where not every run reached the stage. ``Other steps'' is the largest median share among segmentation, statement, and review steps. The static compile and the Lean checks inside the proof loop spend time, not tokens (text). Rows cover the same runs as \Cref{tab:analyst-cost}; the rejected llm.c run (GLM~5.3) stopped in the race-freedom fallback, so it reached no proof stage.}
\label{tab:stage-cost}
\setlength{\tabcolsep}{5pt}
\renewcommand{\arraystretch}{1.15}
\resizebox{\linewidth}{!}{%
\begin{tabular}{@{}l l r r r r c@{}}
\toprule
\textbf{Set} & \textbf{Model} & \textbf{Segmentation proof} $\mathrm{SEG}(K)$ & \textbf{Equivalence proof} $\mathrm{EQ}(K,S)$ & \textbf{Diagnosis} & \textbf{Race-freedom fallback} & \textbf{Other steps} \\
\midrule
\multirow{4}{*}{Level 1 (PTX, CUDA)} & DeepSeek V4.1 Flash & 271.21 (68.7\%) & 164.53 (36.2\%; 15 of 24) & 14.85 (3.6\%) & --- & $\le 0.2\%$ \\
 & GPT-6 Luna & 646.92 (93.6\%) & 267.78 (38.5\%; 9 of 24) & 35.95 (4.4\%) & --- & $\le 0.2\%$ \\
 & GLM 5.3 Flash & 719.16 (55.1\%) & 454.39 (48.3\%; 21 of 24) & 10.34 (0.7\%) & --- & $\le 0.1\%$ \\
 & GLM 5.3 & 116.39 (42.8\%) & 186.60 (54.8\%) & 4.12 (1.3\%) & --- & $\le 0.3\%$ \\
\midrule
\multirow{2}{*}{Level 2 (PTX)} & GLM 5.3 & 103.96 (29.6\%) & 388.21 (72.5\%; 4 of 5) & 3.29 (1.0\%) & --- & $\le 0.2\%$ \\
 & Opus 5.5 & 16.26 (27.2\%) & 60.76 (69.4\%) & 2.71 (1.4\%; 2 of 5) & --- & $\le 1.0\%$ \\
\midrule
\multirow{2}{*}{Production} & GLM 5.3 & 106.04 (47.7\%; 4 of 5) & 137.63 (49.8\%; 4 of 5) & 3.48 (1.3\%; 4 of 5) & 10.30 (100\%; 1 of 5) & $\le 0.8\%$ \\
 & Opus 5.5 & 21.27 (25.6\%) & 56.03 (53.5\%) & 0.76 (0.7\%; 4 of 5) & 48.81 (23.2\%; 3 of 5) & $\le 1.2\%$ \\
\bottomrule
\end{tabular}}
\end{table*}

\subsubsection{Scaling with Kernel Size}\label{app:scaling} Over the 116 PTX and CUDA verifications, the Spearman rank correlation between the compiled program's length in program counters and the run's cost in dollars is 0.61; within a model, between length and tokens, it is 0.43 for DeepSeek~V4.1~Flash, 0.55 for GPT-6~Luna, 0.64 for GLM~5.3~Flash and Opus~5.5, and 0.70 for GLM~5.3. \Cref{fig:scaling} plots the proven runs; for the three models with at least ten of them, a least-squares fit in log coordinates has slope 1.03 (GLM~5.3), 0.90 (GLM~5.3~Flash), and 0.99 (Opus~5.5), so within a model the tokens grow about linearly with the program's length, and the lines differ by the model's token appetite, not by the kernel. Between length and proof turns it is 0.17, rounds 0.20, and between the number of segments and cost 0.17. Proof cost grows with per-block depth, not with the grid.
\begin{figure}
    \centering
    \includegraphics[width=\linewidth]{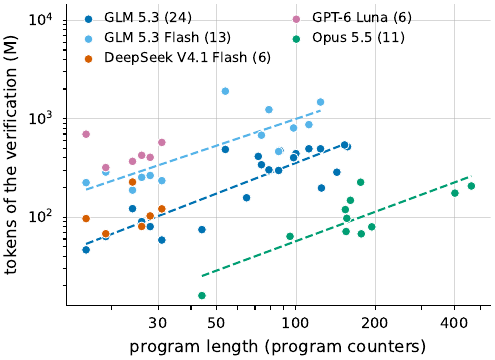}
    \caption{Tokens of each proven verification against the compiled program's length, over the PTX and CUDA runs, one colour per model (the number of runs in parentheses). The dashed lines are least-squares fits in log coordinates for the models with at least ten proven runs, a guide to the eye, not a fitted law; DeepSeek~V4.1~Flash and GPT-6~Luna prove only the elementwise kernels and get none.}
    \label{fig:scaling}
\end{figure}

\subsubsection{Why Every Model Fails on BatchMatmul and DepthwiseConv}\label{app:hard} Read from the 16 failing Level~1 runs on PTX and CUDA (four models, two kernels): no statement is false and no verdict blames the kernel; every analyst verdict but two is ``theorem true, proof hard'' or ``proof loop stuck''. Every model stops on the same proof. Both kernels pin a 128-thread block whose warps run long chains of tensor-core operations, each once per warp: BatchMatmul's 16-trip $K$-loop of \texttt{ldmatrix} and \texttt{mma} makes one warp's run of a block phase up to 97 sub-phases, and DepthwiseConv runs 26 collective operations back to back. The generated scaffolding stops at the lane: the per-lane bridges and the trip lemmas are given, but composing them into the warp-level fold across the loop and the collective steps is left to the agent, and that is where every stall is. The two weaker models never close the first equivalence; the two GLM models close it and run out of budget on the second, whose proof must carry the whole dataflow through the staged shared buffers and the fragment-layout arithmetic. Two runs ended on a genuine bug in GPT-6~Luna's own segmentation, caught by the first equivalence as designed. What would move these kernels is a generated warp-level bridge per block phase, so that the part every model fails on becomes generated, checked text. On HIP the pinned BatchMatmul kernel runs one 64-lane wavefront per block, so the warp-level composition that stops every model here is absent, and GLM~5.3 proves it; DepthwiseConv fails on HIP as on the other two languages.

\subsubsection{Model Size and Tokens}\label{app:model-size} Taking the active parameter count as a proxy for capability, \Cref{fig:model-size} plots, for the three small models with published sizes, the Level~1 pass rate, the input tokens per proved kernel, and the input tokens per kernel against active parameters, over PTX and CUDA. The pass rate rises with model size, and the larger model is also the cheaper one per proof: the pass rate goes from 25\% to 54\% to 79\%, and a model that can close a proof needs fewer turns to do it, while a model that cannot spends its rounds and stops. Input tokens per proved kernel: GLM~5.3 (40B active) 466M, DeepSeek~V4.1~Flash (16B) 1,617M, GLM~5.3~Flash (18B) 2,170M, and GPT-6~Luna 3,313M; mean input tokens per kernel, proved or not: 369M, 404M, 1,175M, and 828M. The largest model is the most efficient per proof, by 3.5 and 4.7 times; DeepSeek and GLM~5.3 spend about the same per kernel, but GLM~5.3 turns it into 19 certificates against 6, while GLM~5.3~Flash fails expensively, running 7 of its 11 failures to the ceiling. Output tokens are 0.7 to 2.4\% of input for every model, so the input count carries the comparison; the dialogue re-sent each turn, mostly as cache reads, is the cost, not what the model writes. In dollars the order of DeepSeek and GLM~5.3 per proof is reversed (\$39 against \$73), only because DeepSeek's tokens are priced lowest. On Level~2 the gap is wider: Opus~5.5 proves every kernel at 107M input tokens per kernel, a quarter of the 374M GLM~5.3 spends per kernel (\Cmain{tab:verify-model}).
\begin{figure}
    \centering
    \begin{subfigure}{\linewidth}
        \centering
        \includegraphics[width=\linewidth]{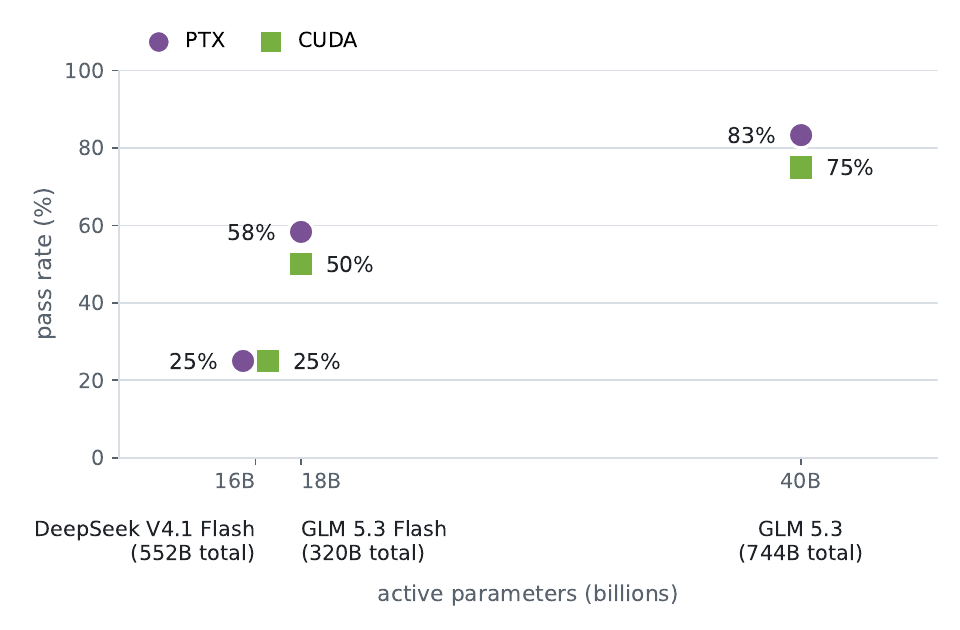}
        \caption{Pass rate.}
        \label{fig:model-size-pass}
    \end{subfigure}
    \begin{subfigure}{\linewidth}
        \centering
        \includegraphics[width=\linewidth]{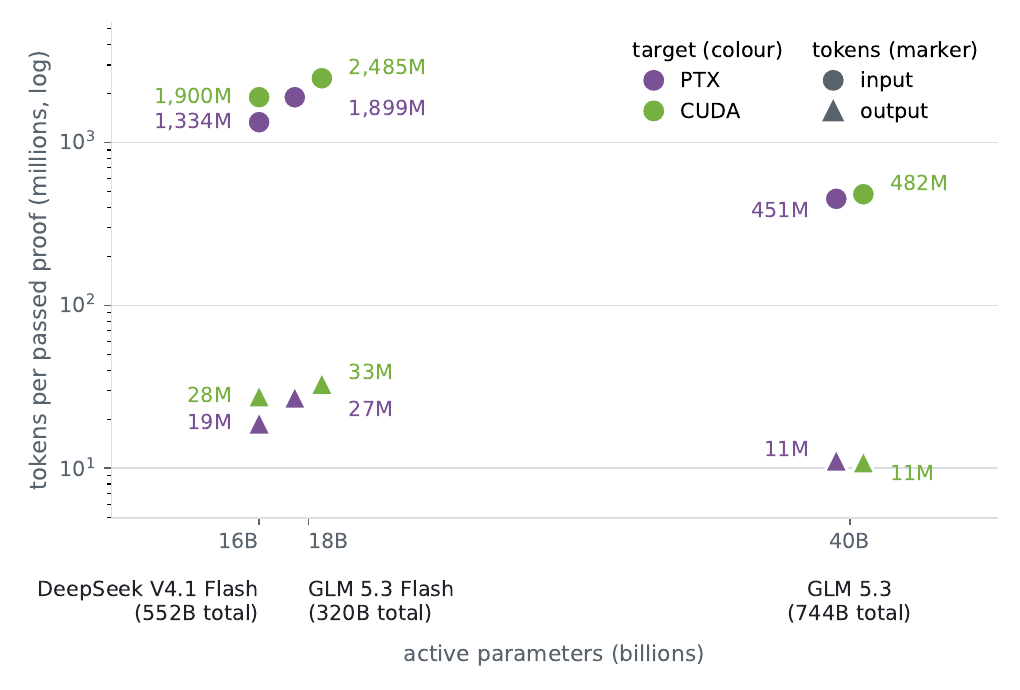}
        \caption{Input and output tokens per proved kernel.}
        \label{fig:model-size-cost-proof}
    \end{subfigure}
    \begin{subfigure}{\linewidth}
        \centering
        \includegraphics[width=\linewidth]{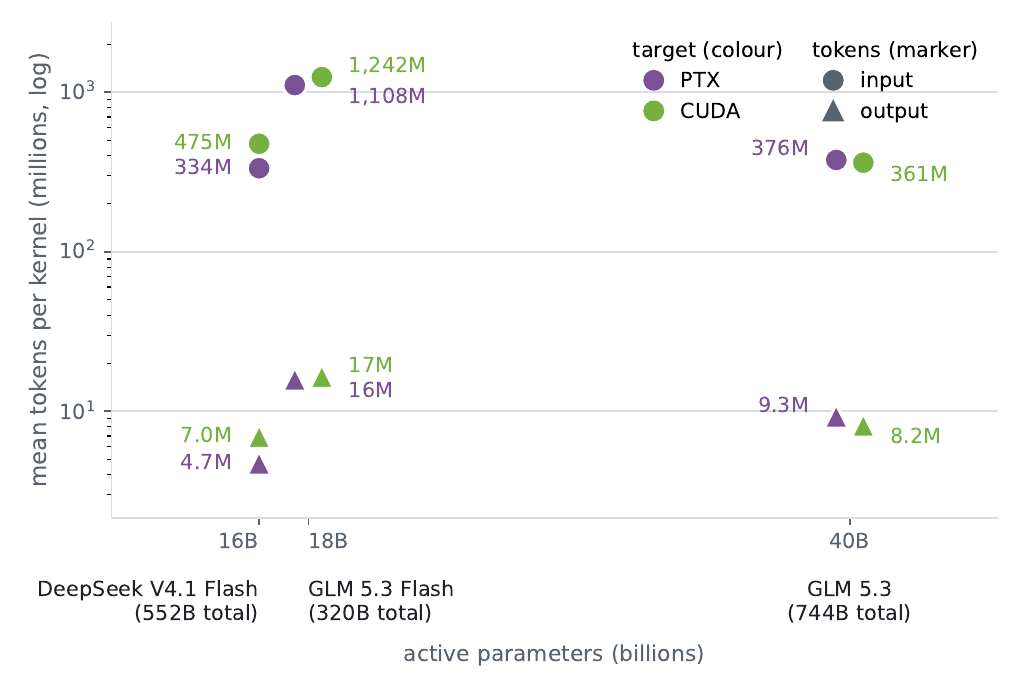}
        \caption{Mean input and output tokens per kernel, proved or not.}
        \label{fig:model-size-cost-kernel}
    \end{subfigure}
    \caption{Level~1 pass rate and tokens against the number of active parameters of the three small models with published sizes, over PTX and CUDA.}
    \label{fig:model-size}
\end{figure}

\subsubsection{The Race-Freedom Fallback}\label{app:fallback} The translator's generated race-freedom lemmas proved on all 40 Level~1 and Level~2 verifications on PTX and CUDA; the agent fallback of \Cmain{sec:advanced} never ran there. On the production kernels it ran on five of ten, where the translator could not name a footprint: \texttt{vllm\_silu\_and\_mul} (8 turns), \texttt{llmc\_softmax\_forward} (10 turns), \texttt{fam\_flash\_attn} (2), \texttt{vllm\_fused\_qknorm\_rope} (7, an output address that divides by the block size), and \texttt{vllm\_awq\_gemm} (6); four were proved, and on \texttt{llmc\_softmax\_forward} the fallback returned a located race instead of a proof (\Capp{app:library}). A later run of the same kernel text reuses the stored proof at no cost.

\subsubsection{The Flat-Model Ablation}\label{app:flat}
Three kernels were proved directly against the flat model (no segmentation, no segmentation proof) and compared with their default runs; tokens are input / output in millions. FlashAttention, Opus~5.5, PTX: flat 222.5 / 3.06, of which the direct proof 219.6 / 2.99 in three rounds of 20 turns, against 172.5 / 2.77 for the default run, of which the equivalence proof 119.6 / 2.05 in two rounds of 11 turns and the segmentation with its proof 49.1 / 0.64. The direct proof used 84\% more input tokens than the equivalence proof and the whole run 29\% more than the default run; it also crossed the \$140 ceiling the run started with and finished only after it was raised to the Level~2 ceiling of \$200, under which the default run had finished. On the two short kernels, ReLU and GEMM with GLM~5.3 on PTX, the direct proof was cheaper than the default run. Caveats: one hard kernel; the model differs between GEMM and FlashAttention; the default runs ran on an earlier code version than the flat runs.

\subsubsection{The Failure Analysis}\label{app:analyst} \Cref{tab:analyst-cost} gives the cost and the verdict statistics of the analysis per model; \Cref{fig:analyst-ratios} the distribution of verdicts by language, model, level, and library.
\begin{table}
\centering
\small
\caption{The failure analysis inside the verification runs, per model: the Level~1 runs on PTX and CUDA, the Level~2 runs, and the production kernels. ``Analysis'' is every analyst, falsity-probe, and second-opinion call. Share: the analysis's input tokens over the run's, summed over a model's runs and as the per-run median. Verdict stops: runs an analyst verdict ended before their spending ceiling. False positives: kernel-bug or race verdicts on a correct kernel, those refuted by the second analyst over those issued; every one was refuted, so none was acted on (two were issued before the second analyst existed and are counted as reviewed and refuted). Refuted: reviews in which the second analyst disputed the verdict, out of those it gave; a disputed verdict is not acted on.}
\label{tab:analyst-cost}
\setlength{\tabcolsep}{4pt}
\renewcommand{\arraystretch}{1.15}
\resizebox{\columnwidth}{!}{%
\begin{tabular}{@{}l c c c c c@{}}
\toprule
\textbf{Model} & \makecell{\textbf{Calls}\\analyst / probe / 2nd op.} & \makecell{\textbf{Share of cell input}\\aggregate / median} & \makecell{\textbf{Verdict}\\\textbf{stops}} & \makecell{\textbf{False positives}\\refuted / issued} & \makecell{\textbf{2nd op.}\\refuted} \\
\midrule
DeepSeek V4.1 Flash & 610 / 755 / 3 & 3.7\% / 3.5\% & 1 & 1 / 1 & 3 / 3 \\
GPT-6 Luna & 599 / 815 / 5 & 4.7\% / 4.3\% & 6 & 1 / 1 & 2 / 5 \\
GLM 5.3 Flash & 336 / 303 / 8 & 0.7\% / 0.6\% & 1 & 1 / 1 & 6 / 8 \\
GLM 5.3 & 154 / 176 / 4 & 1.1\% / 1.1\% & 2 & 0 / 0 & 0 / 4 \\
Opus 5.5 & 3 / 10 / 0 & 0.8\% / 0.5\% & 0 & 0 / 0 & 0 / 0 \\
\bottomrule
\end{tabular}%
}
\end{table}

\begin{figure}
    \centering
    \begin{subfigure}{\linewidth}
        \centering
        \includegraphics[width=\linewidth]{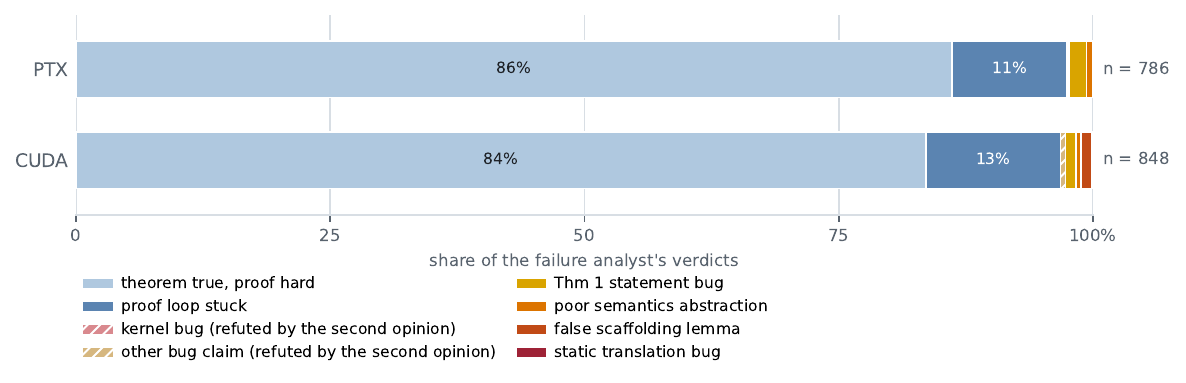}
        \caption{Level~1, by target language.}
        \label{fig:analyst-l1-isa}
    \end{subfigure}\\[4pt]
    \begin{subfigure}{\linewidth}
        \centering
        \includegraphics[width=\linewidth]{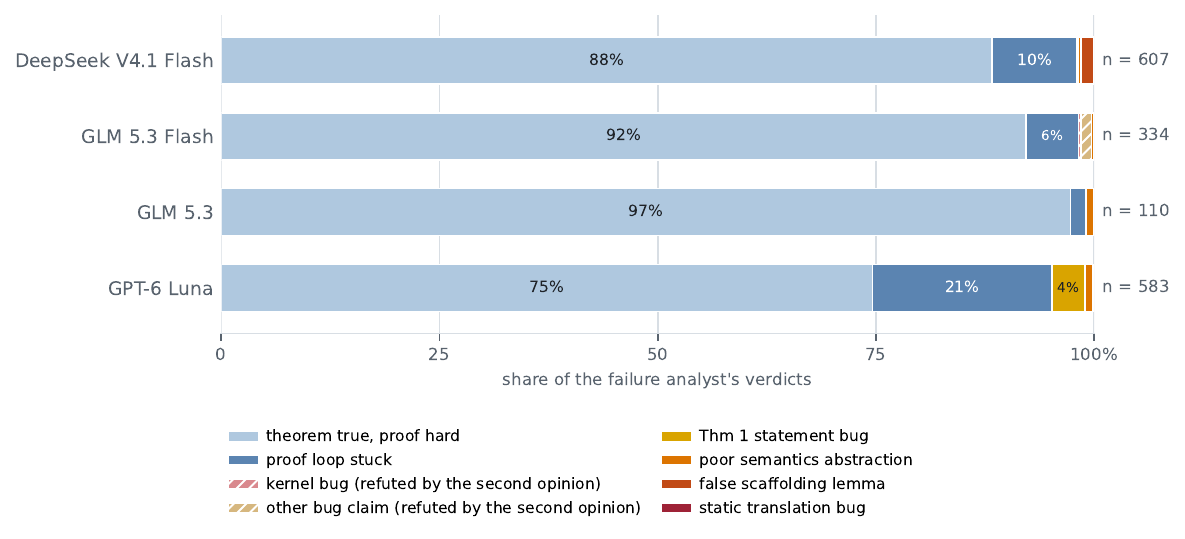}
        \caption{Level~1, by model.}
        \label{fig:analyst-l1-model}
    \end{subfigure}\\[4pt]
    \begin{subfigure}{\linewidth}
        \centering
        \includegraphics[width=\linewidth]{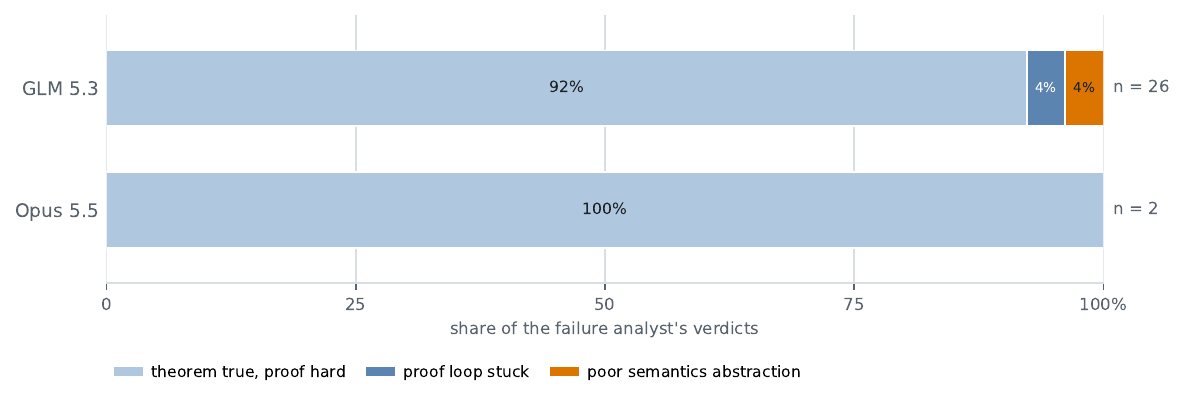}
        \caption{Level~2, by model.}
        \label{fig:analyst-l2-model}
    \end{subfigure}\\[4pt]
    \begin{subfigure}{\linewidth}
        \centering
        \includegraphics[width=\linewidth]{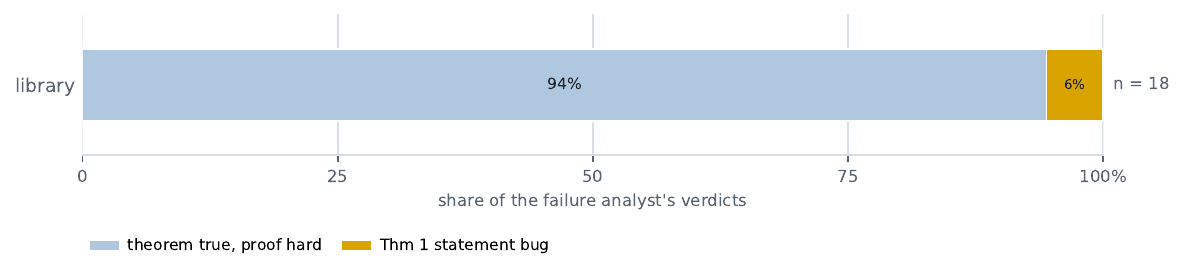}
        \caption{Production kernels.}
        \label{fig:analyst-library}
    \end{subfigure}
    \caption{Verdicts of the failure analysis, as shares of the verdicts given.}
    \label{fig:analyst-ratios}
\end{figure}

\subsubsection{Latency}\label{app:latency} \Cref{tab:latency} and \Cref{fig:latency} give the latency of the verified evaluation kernels against torch eager, \texttt{torch.compile}, and the fastest vendor library, on an H100 for PTX and CUDA and on an MI300X for HIP; every ratio is against baselines measured on the same device, so the targets are not compared with each other. The Level~2 kernels were verified on PTX only, so PTX covers 17 kernels and CUDA and HIP the twelve Level~1 kernels. The verified kernels beat torch eager on every kernel (geometric mean 5.5$\times$ on PTX, 4.2$\times$ on CUDA, 4.1$\times$ on HIP); \texttt{torch.compile} with max-autotune on 14 of 17 on PTX (2.6$\times$), 9 of 12 on CUDA (1.9$\times$), and 10 of 12 on HIP (2.1$\times$); and the fastest vendor library on 11 of the 14 PTX kernels that have one (1.46$\times$), 6 of 9 on CUDA (1.37$\times$), and 7 of 10 on HIP (1.22$\times$). The PTX kernels are faster than the CUDA ones on 6 of the 12 Level~1 kernels, by the most on GEMV (1.8$\times$), BatchMatmul (1.8$\times$), and DepthwiseConv (1.4$\times$); the other six are within 5\%. The losses are the launch-bound elementwise kernels (2 to 5\% behind \texttt{torch.compile} on the H100); \texttt{rmsnorm} against FlashInfer (1.3$\times$) and, on HIP, against \texttt{torch.compile} (1.8$\times$) and AITER (1.5$\times$); \texttt{batchmatmul} against the vendor batched GEMM (cuBLAS, built on wgmma and TMA, which are outside LARK's scope, 1.3$\times$ on PTX and 2.3$\times$ on CUDA; hipBLASLt 2.0$\times$); SwiGLU on PTX against vLLM's fused MLP (1.05$\times$); GEMV on CUDA and HIP against \texttt{torch.compile} (1.9$\times$, 2.0$\times$), where the PTX version wins; and LayerNorm on HIP against AITER (1.2$\times$).
\begin{table*}
\centering
\footnotesize
\caption{Latency of the verified evaluation kernels at their pinned problem sizes (median of 7 rounds): PTX and CUDA on an H100 80GB SXM, HIP on an MI300X. The Level~2 kernels were verified on PTX only. ``Ours'' is the kernel's time in microseconds; a ratio is the baseline's time over ours on the same device, so a value above 1 means the verified kernel is faster, and the targets are not compared with each other. Baselines: torch eager (the task's Torch module), \texttt{torch.compile} with max-autotune, and the fastest vendor library that offers the operation (cuBLAS, cuDNN, vLLM, FlashInfer, and the torch attention backends on the H100; hipBLASLt, rocBLAS, MIOpen, and AITER on the MI300X; a dash where none does).}
\label{tab:latency}
\setlength{\tabcolsep}{3pt}
\renewcommand{\arraystretch}{1.1}
\resizebox{\linewidth}{!}{%
\begin{tabular}{@{}l rrrr rrrr rrrr@{}}
\toprule
& \multicolumn{4}{c}{\textbf{PTX (H100)}} & \multicolumn{4}{c}{\textbf{CUDA (H100)}} & \multicolumn{4}{c@{}}{\textbf{HIP (MI300X)}} \\
\cmidrule(lr){2-5} \cmidrule(lr){6-9} \cmidrule(l){10-13}
\textbf{Kernel} & Ours ($\mu$s) & vs eager & vs compile & vs vendor & Ours ($\mu$s) & vs eager & vs compile & vs vendor & Ours ($\mu$s) & vs eager & vs compile & vs vendor \\
\midrule
\texttt{relu} & 2.05 & 3.53 & 0.96 & --- & 2.01 & 3.60 & 0.98 & --- & 1.73 & 4.35 & 1.07 & --- \\
\texttt{silu} & 2.08 & 3.55 & 0.97 & --- & 2.06 & 3.58 & 0.98 & --- & 1.75 & 3.81 & 1.10 & --- \\
\texttt{gelu} & 1.99 & 3.52 & 1.02 & 2.38 & 1.98 & 3.54 & 1.03 & 2.39 & 1.77 & 4.17 & 3.02 & 1.72 \\
\texttt{softmax} & 4.12 & 2.93 & 3.06 & 3.00 & 4.37 & 2.76 & 2.88 & 2.83 & 5.25 & 2.23 & 2.19 & 1.32 \\
\texttt{layernorm} & 5.03 & 6.63 & 1.78 & --- & 4.78 & 6.98 & 1.87 & --- & 5.05 & 3.79 & 1.69 & 0.85 \\
\texttt{rmsnorm} & 4.38 & 6.12 & 1.73 & 0.75 & 4.29 & 6.26 & 1.77 & 0.77 & 4.02 & 5.61 & 0.54 & 0.67 \\
\texttt{gemv} & 3.05 & 4.23 & 0.95 & 1.53 & 5.54 & 2.33 & 0.53 & 0.84 & 3.83 & 3.09 & 0.49 & 1.31 \\
\texttt{gemm} & 2.90 & 5.79 & 2.03 & 1.18 & 3.39 & 4.95 & 1.74 & 1.01 & 2.14 & 5.05 & 6.18 & 1.27 \\
\texttt{linear} & 2.88 & 7.40 & 5.93 & 1.26 & 3.62 & 5.88 & 4.71 & 1.00 & 3.36 & 3.74 & 4.34 & 1.02 \\
\texttt{batchmatmul} & 7.24 & 5.66 & 3.96 & 0.77 & 12.86 & 3.19 & 2.23 & 0.43 & 11.13 & 2.90 & 4.08 & 0.51 \\
\texttt{conv2d} & 3.28 & 8.41 & 5.96 & 4.95 & 3.23 & 8.56 & 6.07 & 5.04 & 2.65 & 9.55 & 4.54 & 3.16 \\
\texttt{depthwise\_conv} & 3.36 & 4.06 & 6.12 & 2.38 & 4.65 & 2.94 & 4.43 & 1.72 & 4.13 & 3.83 & 5.29 & 2.00 \\
\texttt{rope} & 2.24 & 20.12 & 1.10 & 1.16 & --- & --- & --- & --- & --- & --- & --- & --- \\
\texttt{ffn} & 9.75 & 4.96 & 4.53 & 1.09 & --- & --- & --- & --- & --- & --- & --- & --- \\
\texttt{swiglu} & 11.03 & 4.99 & 6.11 & 0.95 & --- & --- & --- & --- & --- & --- & --- & --- \\
\texttt{flash\_attn} & 6.01 & 5.13 & 4.77 & 1.25 & --- & --- & --- & --- & --- & --- & --- & --- \\
\texttt{gqa} & 4.32 & 8.10 & 5.57 & 1.22 & --- & --- & --- & --- & --- & --- & --- & --- \\
\midrule
Geometric mean & & 5.50 & 2.61 & 1.46 & & 4.20 & 1.91 & 1.37 & & 4.06 & 2.14 & 1.22 \\
\bottomrule
\end{tabular}}
\end{table*}

\begin{figure*}
    \centering
    \includegraphics[width=\textwidth]{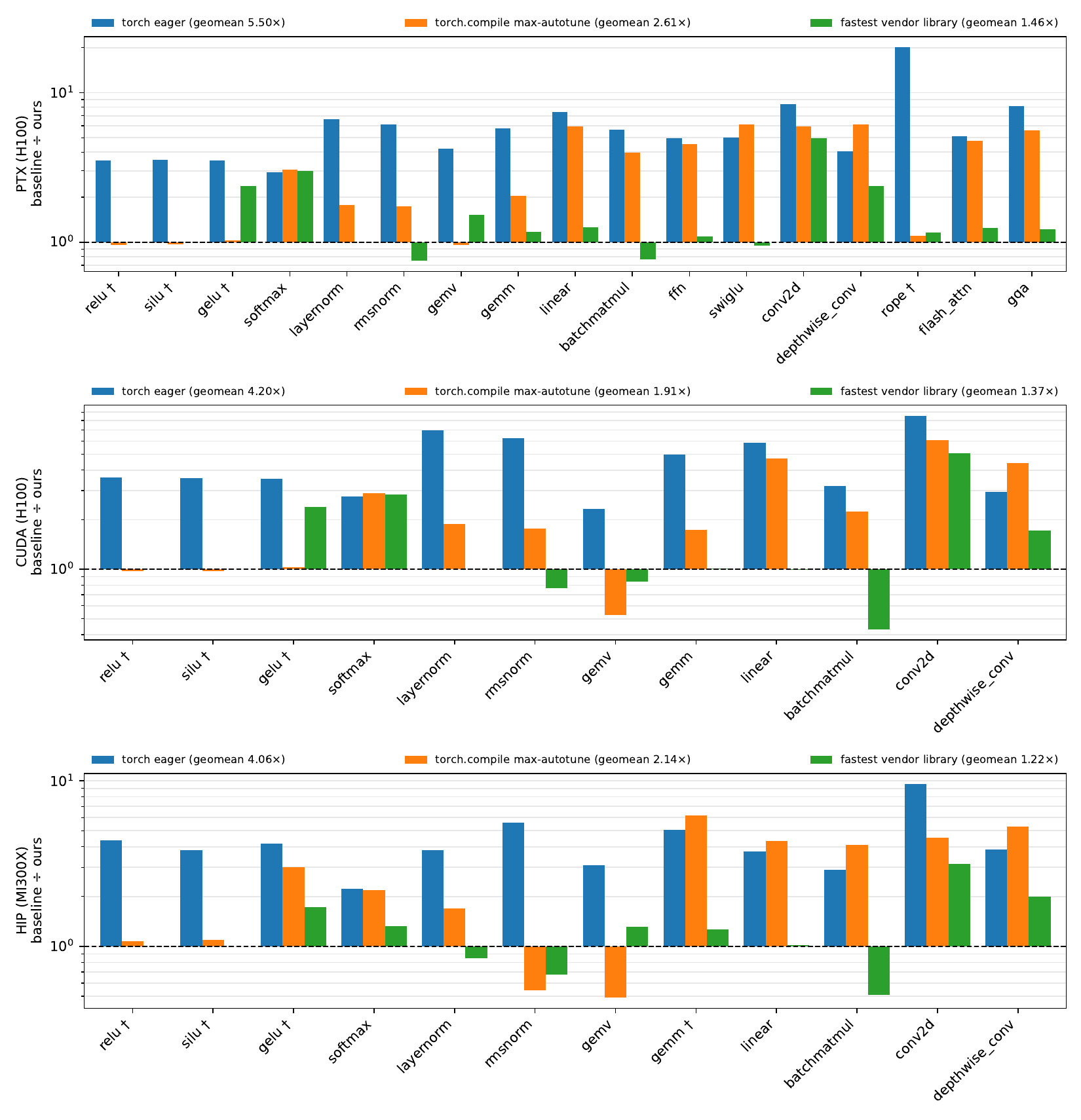}
    \caption{Latency of the verified evaluation kernels relative to the baselines, from the same measurement as \Cref{tab:latency}: PTX and CUDA on the H100, HIP on the MI300X. The vendor libraries are cuBLAS and cuBLASLt (GEMM, GEMV, batched GEMM, linear), cuDNN (convolution), FlashInfer (RMSNorm, softmax, RoPE, prefill attention), vLLM (SiLU-and-mul, GELU, FlashAttention-2), and the torch attention backends on the H100, and hipBLASLt, rocBLAS, MIOpen, and AITER on the MI300X; ``fastest'' is the best of these per kernel. A dagger marks a kernel whose time is within the launch overhead.}
    \label{fig:latency}
\end{figure*}

\subsubsection{Mutation Results Analysis}\label{app:mutation_results} 
\begin{table*}
\centering
\small
\caption{Mutation study: how each non-equivalent mutant was rejected. One mutant per rule,
each counted once by its final run; ReLU has no barrier, so R3 does not apply to it.}
\label{tab:mutation-channels}
\renewcommand{\arraystretch}{1.1}
\begin{tabular}{@{}llcccccc@{}}
\toprule
Kernel & Lang. & Mutants$^a$ & Static$^b$ & Race-free$^c$ & Kernel bug$^d$ & Premise$^e$ & Escaped$^f$ \\
\midrule
ReLU    & CUDA & 4 & 1 & 0 & 2 & 1 & 0 \\
ReLU    & PTX  & 4 & 0 & 1 & 2 & 1 & 0 \\
ReLU    & HIP  & 4 & 0 & 1 & 3 & 0 & 0 \\
\addlinespace
Softmax & CUDA & 5 & 1 & 0 & 4 & 0 & 0 \\
Softmax & PTX  & 5 & 1 & 0 & 4 & 0 & 0 \\
Softmax & HIP  & 5 & 2 & 2 & 0 & 1 & 0 \\
\addlinespace
GEMM    & CUDA & 5 & 4 & 1 & 0 & 0 & 0 \\
GEMM    & PTX  & 5 & 1 & 2 & 2 & 0 & 0 \\
GEMM    & HIP  & 5 & 1 & 3 & 1 & 0 & 0 \\
\bottomrule
\end{tabular}

\vspace{4pt}
\begin{minipage}{0.95\linewidth}
\footnotesize
$^a$\emph{Mutants}: non-equivalent mutants, one per mutation rule (arithmetic, load/store,
barrier, register/variable, loop range). A mutant that computes the same function as the
original is excluded and replaced by the next candidate of its rule. Each row's five outcome
columns sum to this number.\\
$^b$\emph{Static}: rejected by the static compiler before any LLM call: the lane-safety check
(every lane of a warp stores to one cell), a race located at compile time, or a race-freedom
pair lemma that Lean refutes with a counterexample.\\
$^c$\emph{Race-free}: the generated race-freedom proof obligation is judged false, with two
concrete threads and a memory cell; confirmed by an independent second opinion.\\
$^d$\emph{Kernel bug}: the equivalence proof fails and the failure analyst shows the kernel
computes a different function from the specification; confirmed by an independent second
opinion.\\
$^e$\emph{Premise}: the equivalence statement, generated from the kernel's signature and declared
launch, is rejected by Lean, because the mutant needs a premise the standard signature does not
give (e.g.\ an out-of-bounds access).\\
$^f$\emph{Escaped}: the mutant was proved equivalent to the specification, i.e.\ not detected.
\end{minipage}
\end{table*}

\Cref{tab:mutation-channels} reports final-run outcomes after corrections to specification premises and output extraction: 11 static rejections, 10 race-freedom failures, 18 kernel-bug diagnoses, and 3 premise rejections; no mutant escapes. Static and race-freedom checks account for 12 of 15 GEMM rejections, while equivalence failure analysis accounts for most ReLU and Softmax rejections.

\subsection{Measuring the Per-Operation Semantics against the Hardware}
\label{app:non-ctx-sem-fuzzing}

Every lower-layer definition of \Cmain{sec:foundations} is checked against the GPU before it is pinned. For each operation in a language's table, a harness draws random operands from declared strata (finite values of each width, $-\infty$ where the operation admits it, boundary cases such as zero and powers of two), runs the operation on the GPU in its own operand form, and compares the result with the Lean definition evaluated on the same operands; a disagreement outside the declared idealizations fails the sweep. The sweep makes no LLM calls and is re-run whenever a table changes; its record is pinned with the table, and a run reads the record its table was pinned on, never the newest. The sweep is what turned one wrong definition into a right one: the approximate base-two exponential had been formalized as $e^x$ and is $2^x$. Shuffles and vector accesses are measured per lane and per alignment in the same way, and the lane layouts of the warp-collective operations are recorded from the hardware before they are pinned.

\subsection{Retargeting LARK to a New Target Language}
\label{app:retarget}

\paragraph{Effort.} CUDA support on top of PTX took at most 39 person-days; its written lines (code, tests, and documentation, generated tables apart) are 68,741, of which 35,314 are code. HIP, ported from CUDA, took 21,495 written lines, of which 15,333 are code, about a third of CUDA's lines and under half of its code. The shared machine model changed by 885 lines added and 655 removed, including one generalization: the warp width became a parameter, 32 for PTX and CUDA and 64 for HIP, whose wavefront has 64 lanes. HIP's own translator code is largely a copy of CUDA's, with 91\% of its lines identical. CUDA's own code is 37 files and 14,062 Python lines and HIP's 30 files and 7,535 lines, against 221 files and 127,988 lines shared across languages; of the machine model's Lean text, 96\% is shared for HIP, 94\% for CUDA, and 86\% for PTX.

\paragraph{The split in symbols.} What is fixed once splits as $M = M_{\mathrm{launch}} \cup L$. The shared part
\[
  M_{\mathrm{launch}} = (\Sigma, \mathrm{run}, \mathrm{conv}, \mathrm{RF}, \mathrm{Pre}_M, \mathrm{Ext}_M, \mathcal{C})
\]
describes a launch for every language: the state, the folds of \Cmain{eq:fold}, race freedom $\mathrm{RF}$ and its library (\Cref{sec:advanced}), the premise and extractor definitions $\mathrm{Pre}_M$ and $\mathrm{Ext}_M$ (\Cref{sec:bridge}), and the certificate $\mathcal{C}$ (\Cref{sec:trust}). The per-language part
\[
  L = (\mathcal{I}, \mathrm{sem}, \mathrm{cut}, w_L, \mathrm{parse}_L, \mathrm{fmt}_L, \mathit{layout}_L, \mathrm{lower}_L)
\]
is the operation table with its lower-layer definitions, the warp width $w_L$ (32 for PTX and CUDA, 64 for HIP), the reader and the read-back formatter, the lane layouts of the warp-collective operations, and, for a structured language, the lowering.

\Cref{tab:retarget} lists the components of LARK by whether they are shared across target languages or written per language. The split follows one rule: what describes a parallel launch is written once, what describes a language's operations is written per language. The machine model $M$ abstracts GPU execution in general, so its state, value representation, cut hierarchy, folds, and convergence rule, the definition of race freedom with the library that discharges it, the premise schemas, the certificate structure with its end-to-end theorem, the premise and extractor definitions over a signature, the agent loop, and the report are one Lean text and one harness shared by every language. A language supplies its operation set $\mathcal{I}$, generated from a pinned table whose lower-layer definitions are measured against the hardware (\Capp{app:non-ctx-sem-fuzzing}); its semantics $\mathrm{sem}$ and cut classification $\mathrm{cut}$; a parser and printer, so that the byte round trip of \Cmain{sec:foundations} can check every reading; the lane layouts of its warp-collective operations, recorded from the hardware; and, for a structured language, the lowering of source to the operation sequence, which is itself a pinned Lean definition. The translator's race-freedom layer is one algorithm with a per-language front end that says what an operation's operands and memory window are. PTX, being unstructured by design, needs no lowering; CUDA and HIP share the structured-tree lowering and differ in their operation tables and collective layouts.

\begin{table*}
  \centering
  \footnotesize
  \renewcommand{\arraystretch}{1.15}
  \caption{Components of LARK by whether they are shared across target languages or written per language, as instantiated for PTX, CUDA, and HIP. \emph{Shared} components are one artifact used verbatim by every language; \emph{shared core} components have one algorithm and one emitted Lean text with a thin per-language front end; \emph{per language} components are written for each target and, where marked, measured against its hardware.}
  \label{tab:retarget}
  \resizebox{\linewidth}{!}{%
  \begin{tabular}{l p{6.6cm} l p{5.0cm}}
    \toprule
    \textbf{Component} & \textbf{What it contains} & \textbf{Status} & \textbf{Per-language part} \\
    \midrule
    \multicolumn{4}{l}{\emph{Machine model $M$}} \\
    State and memory hierarchy & One state for the launch: registers keyed by thread and name, global, shared, and local memory with type tags, coordinates and geometry & Shared & Coordinate reads and address spellings \\
    Value representation & Typed values: exact reals with $-\infty$, exact integers with a width tag, predicates, packed halves; partial operations (faults) & Shared & \\
    Cut hierarchy and folds & $\mathrm{cut}$ kinds, the folds of \Cmain{eq:fold}, convergence, fuels & Shared & Which operations define which cut; the warp width (32 or 64) \\
    Race freedom & Definition over footprints, span lemmas, pair-lemma library, the reduction metatheory relied on & Shared & \\
    Premise schemas & Placement, separation, parameters, alignment, geometry, fuel, entry (\Capp{app:schemas}) & Shared & \\
    Certificate & The certificate structure, non-vacuity, the end-to-end theorem proved once & Shared & \\
    Operation set $\mathcal{I}$ and $\mathrm{sem}$ & Operation type generated from a pinned table; lower-layer definitions; contextual semantics & Per language & Measured against the hardware (\Capp{app:non-ctx-sem-fuzzing}) \\
    Collective layouts & Lane and fragment layouts of \texttt{ldmatrix}, \texttt{mma}, \texttt{wmma}, shuffles & Per language & Recorded from the hardware \\
    Source lowering & Structured source tree to operation sequence, as a pinned Lean definition & Per language & PTX: none; CUDA and HIP: shared tree lowering \\
    Torch module formalization & Lean definition of the module, tested against the module & Per module & Shared by every language \\
    \addlinespace
    \multicolumn{4}{l}{\emph{Translator}} \\
    Parser and printer & Source to data and back; byte round trip & Per language & \\
    Refusals & Closed list of named refusals (\Capp{app:unsupported}) & Shared mechanism & Per-language spellings \\
    Grid-barrier recognition & Pinned template (PTX) or the one-line call (CUDA, HIP) & Per language & \\
    Race-freedom layer & Footprint regions, pair lemmas and their proofs, agent fallback & Shared core & Operand and memory-window reader \\
    Scaffolding & Phase structure, counted loops and bridges, fuels & Shared core & Loop and branch spellings \\
    \addlinespace
    \multicolumn{4}{l}{\emph{Proof generation and reporting}} \\
    Statement generation & The fixed premise and extractor over a signature; non-vacuity proved once & Shared & \\
    Agent loop and failure analysis & Segmentation, proofs, diagnosis & Shared & \\
    Report and export & Trusted-assumption report, standalone Lean project & Shared & \\
    \bottomrule
  \end{tabular}}
\end{table*}

\subsection{Harness Details}
\label{app:harness-details}

\begin{table*}
  \centering
  \footnotesize
  \renewcommand{\arraystretch}{1.15}
  \caption{Basis for acceptance at each stage of LARK. \emph{Verified}: a Lean proof for the kernel at hand. \emph{Trusted}: pre-checked or declared, kernel-independent, listed in the trusted-assumption report (\Cmain{sec:trust}). \emph{Judged}: data a run supplies, quoted in the report. \emph{Untrusted}: agent-generated, covered by a later Verified row. \Cmain{fig:flowchart-compare} uses the same statuses.}
  \label{tab:trust}
  \resizebox{\linewidth}{!}{%
  \begin{tabular}{l l p{6.2cm} p{1.6cm} p{4.4cm}}
    \toprule
    \textbf{Stage / artifact} & \textbf{Produced by} & \textbf{Basis for acceptance} & \textbf{Status} & \textbf{If it fails} \\
    \midrule
    \multicolumn{5}{l}{\emph{Offline, once per target language or per module}} \\
    Machine model $M$ & Humans, agents & Review; pinned by hash; its idealizations declared as scope & Trusted & (pre-checked) \\
    Per-operation semantics & Generated from pinned tables & Measured against the hardware; pinned by hash & Trusted & Operation not in table: refuse \\
    Torch module formalization & Agents, humans & Tested against the Torch module on random inputs & Trusted (tested) & Mismatch: fix before any run \\
    Premise and extractor definitions & Pinned in $M$ & One definition each over a signature; pre-checked & Trusted & (pre-checked) \\
    Reduction metatheory & Literature & Race-free phases $\Rightarrow$ one final state for every interleaving & Trusted & Reported \\
    \addlinespace
    \multicolumn{5}{l}{\emph{Per kernel}} \\
    Kernel as data & Translator & Read-back check: the data formatted back to text equals the source byte for byte & Verified (read-back) & Refuse \\
    Flat model & Derived in $M$ & Built from the data by the pinned definition & Verified (by construction) & \\
    Race-freedom obligation $\mathrm{RF}(K)$ & Translator; agent fallback & Lean proof per phase and pair of units & Verified & Reject \\
    Segment-level model & Agent & None; covered by $\mathrm{SEG}(K)$ & Untrusted & See $\mathrm{SEG}(K)$ \\
    Theorem statement & Generated from the signature & The fixed premise and extractor over the task's signature, which the task's tests launch with & Trusted (signature) & Entry point mismatch: refuse \\
    Non-vacuity & Proved once in $M$ & The fixed premise satisfiable for every signature and input & Verified & \\
    Segmentation obligation $\mathrm{SEG}(K)$: flat $=$ segment-level & Agent & Lean proof; no \texttt{sorry}; standard axioms & Verified & Retry; then no certificate \\
    Equivalence obligation $\mathrm{EQ}(K,S)$: segment-level $=$ specification & Agent & Lean proof; no \texttt{sorry}; standard axioms & Verified & Retry; then no certificate \\
    End-to-end theorem & Proved once in $M$ & Assembles the rows above into \Cmain{eq:equiv}; \texttt{\#print axioms} & Verified & \\
    Generator hints & Kernel generator & Regions: checked by the lemmas; constant marks and launch: reported & Untrusted / Judged & Failed proof / reported \\
    \addlinespace
    \multicolumn{5}{l}{\emph{Standing}} \\
    Named constants and approximate instructions & Table & Read as exact values & Trusted & Reported \\
    Grid-barrier template & Pinned & A barrier under a cooperative launch & Trusted & Other spelling: refuse \\
    Read-back formatter & Fixed per language & Formats the kernel data back to text; about a quarter the size of the reader & Trusted & Reported \\
    Floating-point behavior & Numeric test & Outside the theorem; tested at a tolerance & Scope & Reported \\
    Lean, Mathlib, axioms & External & \texttt{\#print axioms} per theorem & Trusted & Reported \\
    \bottomrule
  \end{tabular}}
\end{table*}

\paragraph{Grid barrier.} Block-level barriers are single operations with hardware-defined semantics and a cut of their own. A grid-wide barrier is not: in PTX, \texttt{grid.sync()} lowers to atomic counters and spin loops. LARK pins one template of that lowering, which a PTX kernel copies verbatim with its own registers and labels; the translator reads the copy as one grid-barrier pseudo-operation whose body is the identity and whose cut is grid-level, and the printer re-emits the template from the names it read, so the recognition is covered by the byte round trip. What remains trusted is that the template is a barrier under a cooperative launch. In CUDA and HIP the barrier is the one-line \texttt{this\_grid().sync()}. Any other spelling is refused.

\paragraph{Value types.}\label{app:values} The type tags of \Cmain{sec:foundations} are $\mathit{Ty} = \{\mathrm{f16}, \mathrm{f32}, \mathrm{f16x2}, \mathrm{int}_w, \mathrm{pred}\}$: a 16- or 32-bit float with payload in $\mathbb{R} \cup \{-\infty\}$, two packed 16-bit floats held as one value, an integer of width $w$ with payload in $\mathbb{Z}$, and a predicate. Other formats (bf16, 64-bit floats, packed 8-bit types) are outside the set and refused by name (\Capp{app:unsupported}). The payload of a float tag is an element of $\mathbb{R} \cup \{-\infty\}$, with $-\infty$ admitted because kernels use it as the initial value of a running maximum; of a packed tag a tuple of them; of an integer tag an element of $\mathbb{Z}$, the width being part of the tag; of a predicate an element of $\{0,1\}$. Rounding is hard to reason about, which is why arithmetic is exact in $M$ and floating-point behavior is left to testing (\Cmain{sec:trust}).

\paragraph{Read-back check.}\label{app:readback} The translator's reading of a kernel is not verified; each of its outputs is checked instead, in the manner of translation validation~\citep{translationvalidation,translationval4oc}: the parsed data is formatted back to text and compared byte for byte with the source, so a wrong reading passes only if the formatter has a compensating defect. The trust therefore moves from the reader to the read-back formatter, which is the smaller and simpler component: for PTX the reader is 1,405 lines against 320 for the formatter, for CUDA 2,313 against 570, with a 217-line helper shared by both. The formatter walks the emitted data, not the reader's in-memory objects, so what is printed is the data the theorem is about. The formatter is a trusted item and is listed in the report.

\paragraph{Fuel.} Each level of the fold of \Cmain{eq:fold} runs under a fuel: lane steps per sub-phase, sub-phases per block-level phase, block-level phases per grid-level phase, and grid-level phases per launch. The fuels are data a run supplies, and exhausting any of them is a fault, so a fuel too small for the kernel fails the proof rather than weakening the theorem; nothing about loop bounds is trusted or analysed by a trusted component. The translator proposes generous fuels from the source, and the proof carries termination.

\paragraph{Folds and the proof-friendly restatement.} The folds of \Cmain{eq:fold} are pinned once in $M$. The agent's segment-level model restates the kernel's run in a sequential shape that a proof can induct on; the first equivalence of \Cmain{sec:bridge} proves that this restatement equals the fold, using the verified race freedom of the kernel, and the second relates the restatement to the specification.

\paragraph{Efficiency mechanisms.} Several mechanisms keep the agent--checker loop efficient: the agent can test lemmas in scratch Lean scripts before committing to them; a timed-out check returns the slowest declarations, the lemmas that passed, and any errors raised, so a failure still yields information; \textsc{progress} and \textsc{delta} submissions let the agent resume from a partial proof or patch part of a program instead of re-emitting it; and context compaction drops the oldest dispensable turns and repeated listings while keeping the system prompt, the first turn, and the most recent messages.

\paragraph{Failure analysis.} When a loop exhausts its attempts, an agentic failure analysis diagnoses the cause (a hard proof, an inefficient proving approach or formalization, or a faulty kernel) and acts accordingly: continue the loop, restart it with a fresh dialogue, redo the formalization, or return the diagnosis to the generation side as feedback for a new kernel. The last path closes the generate-and-verify cycle: a kernel that cannot be proved equivalent because it is wrong is sent back with an explanation of why. The pipeline aborts only when the number of analyses or the overall budget reaches its ceiling.

\subsection{Premise Schemas}
\label{app:schemas}

Every theorem statement is computed from the task's \emph{signature}: a list with one entry per kernel argument, in entry order. A buffer entry carries its role (input, output, in-place, or scratch), its element type, its shape, and its layout, a strided tensor in PyTorch's sense, with each stride a literal or a named integer parameter; a scalar entry carries its domain (any real, or positive) or its pinned value; an integer dimension carries its value. The launch is recorded with it: the block pinned or capped at the declared bound, and the grid pinned, capped and one-dimensional, or merely non-empty, the form chosen by data. The signature is generated from the task's registry, the same list the task's tests launch the kernel with, and is checked against the kernel's entry point by position and kind. The fixed premise over a signature states: every input and in-place buffer placed at its base under its layout, with the memory's type tags set; every pair of distinct buffers byte-separated; every base aligned to its element size and vector width; every scalar bound to the specification's input, or pinned; every dimension pinned; every runtime stride at least its row length; and the launch geometry. A scratch buffer is separated and aligned, never placed or extracted; an in-place buffer is placed as an input and read as an output. The fixed extractor reads every output and in-place buffer under its layout. Both are defined once in $M$; the facts a proof needs (a placement, a separation, a binding) are projection lemmas of the model, instantiated by name for the run. Satisfiability of the premise is one theorem of $M$ over every signature whose byte spans fit. Layouts out of scope, refused by name: sub-byte and bit-packed elements, indirection through index buffers, and swizzled or blocked global layouts; shared-memory layouts are internal to the kernel and never part of the premise.

Formally, the signature is a list $\mathit{sig} \in \mathit{List}(\mathit{Arg})$ with one entry per kernel argument,
\[
\begin{aligned}
  \mathit{Arg} ::={}& \mathrm{buf}(\rho, \tau, \mathit{shape}, \mathit{layout}) \\
  & \mid \mathrm{scalar}(\mathit{dom}) \mid \mathrm{scalar}(v) \mid \mathrm{dim}(n)
\end{aligned}
\]
a buffer with role $\rho \in \{\mathrm{in}, \mathrm{out}, \mathrm{inout}, \mathrm{scratch}\}$, element type $\tau$, shape, and strided layout (element $\vec i$ at $\mathit{base} + |\tau| \sum_k i_k s_k$); a float scalar bound to a specification input of domain $\mathit{dom}$ or pinned to $v$; an integer dimension pinned to $n$. The theorem of \Cmain{sec:bridge} quantifies over the kernel's parameters $\vec p$ (base addresses, strides, scalars; \Cmain{eq:equiv} folds them into $\sigma$), every state satisfying the premise, and every input, and the satisfiability theorem above reads
\[
  \forall \mathit{sig}.\ \forall x \in D.\ \exists \vec p\; \sigma.\ \mathrm{Pre}_M(\mathit{sig}, \mathit{launch})(\vec p, \sigma, x),
\]
proved once for every signature, so no statement holds vacuously.

\subsection{Unsupported Constructs}
\label{app:unsupported}

The translator refuses, by name and before any proof is attempted: $+\infty$ as a value and NaN in any form (the creation of either is a fault); integer wraparound and integer division or remainder by zero; bit operations other than a literal mask, a literal shift, and the full warp mask; float-to-integer conversions out of range; the \texttt{.sat}, \texttt{.ftz}, and bf16 forms, FP64, and 8-bit types; asynchronous operations (\texttt{cp.async}, \texttt{mbarrier}, TMA, \texttt{wgmma}, \texttt{tcgen05}, cluster barriers); atomics and fences used for communication; partially masked barriers, shuffles, and collectives; a grid barrier spelled other than the pinned template; addresses that are not polynomial in the thread coordinates, loop counters, and parameters; blocks with more than one dimension and grids with more than two; and any operation absent from the pinned table. Approximate instructions (\texttt{ex2.approx}, \texttt{lg2.approx}, \texttt{rcp.approx}, \texttt{rsqrt.approx}) are accepted and read as the exact function, and the tf32 form of \texttt{mma} is accepted as an FP32 carrier; both are declared idealizations listed in the report. The complete list, with the name of each refusal, ships with the code.

\subsection{Race-Freedom Lemmas}
\label{app:racefree}

Race freedom is what the fold of \Cmain{eq:fold} needs because
\begin{multline*}
      \mathrm{RF}(K) \wedge \mathrm{run}^{\vec\kappa}_{\mathrm{grid}}(\sigma) = \sigma' \\ \Longrightarrow \sem{K}_M(\sigma){\downarrow} \wedge \sem{K}_M(\sigma) = \sigma',
\end{multline*}
that is, every execution of $K$ from $\sigma$, under any scheduling, is finite and ends in $\sigma'$: the actions of race-free units commute, and the barrier that ends a phase orders the phases (Lipton's reduction, \Cmain{sec:advanced}).
The translator derives each footprint $W_\phi(u)$, $R_\phi(u)$ of \Cmain{sec:advanced} as a region polynomial in the thread coordinates, loop counters, and kernel parameters, and generates the two disjointness lemmas of the race-freedom obligation for every phase and pair of units, with proofs over the theorem's premises, which Lean checks. Where an address is not polynomial in these variables (a clamped index, for instance), the translator cannot name the region: it emits the lemma's statement without a proof and an agent attempts it, with an independent second opinion on a verdict that the statement is false. A kernel with an unproved lemma is rejected, whether from a real race or a region the translator could not name, and a located race is returned as a counterexample naming the two threads and the cell (\Cmain{sec:experiments}). The condition is sufficient, not necessary: it also rejects units that communicate within a phase, such as producer--consumer blocks.

\subsection{Generator Hints}
\label{app:hints}

A kernel generator may annotate its kernel with three kinds of hint: the regions each thread writes in a phase, which guide the translator's choice of footprints when it generates the race-freedom lemmas; a mark naming the exact constant behind a rounded literal, which the translator cross-checks against the literal's bits and which the pinned decoder then reads as the exact value; and the launch it expects, which selects the launch form of the fixed premise and which the numeric test launches with. The first kind decides nothing, since every lemma built from it is checked by Lean and a wrong region fails a proof. The second and third are the two hints Lean cannot check; both appear in the report.

\subsection{Certificate Export}
\label{app:export}

Every run exports its certificate as a standalone Lean project: the pinned model text, the kernel as data with its parse, the signature, the fuels, the race-freedom lemmas, the two equivalence proofs, and the end-to-end theorem, together with the output of \texttt{\#print axioms} on that theorem. The project builds with the public Lean toolchain and Mathlib and re-checks without any part of LARK, so a certificate can be audited by a checker that shares no code with the one that produced it. The supplementary material contains the project of the GQA kernel.

\subsection{The Kernel-Independent Premises}
\label{app:trusted-base}

The kernel-independent premises $\Pi_M$ of \Cmain{sec:trust} are: (i) the idealizations of $M$, exact arithmetic as scope and the fold of \Cref{eq:fold} with the reduction result of \Cref{sec:advanced}; (ii) the per-operation semantics $\mathrm{sem}$, generated from the pinned tables and measured against the hardware (\Capp{app:non-ctx-sem-fuzzing}); (iii) the pinned grid-barrier template;\label{asm:grid} (iv) a table of four named constants, under which a rounded literal a kernel writes for $\log_2 e$ is read as the exact value, and the approximate instructions read as exact; (v) the Lean formalization of the Torch module; (vi) the premise and extractor definitions $\mathrm{Pre}_M$ and $\mathrm{Ext}_M$; (vii) the read-back formatter (\Capp{app:readback}); (viii) Lean, its three standard axioms, and its toolchain. Item (v) is tested: a Lean metaprogram evaluates the elaborated specification in 64-bit floating point on random inputs against the Torch module, refusing by name any construct it does not recognize, and the result is reported as evidence, not proof.

\subsection{Production Kernels}
\label{app:library}

\paragraph{What ``adapted'' means.} No production kernel is verified byte for byte as upstream ships it. Each adapted kernel is the smallest mechanical rewrite of the upstream kernel into the pipeline's input form, and every step is recorded as a change with its reason. The rewrites common to most kernels change the form of the code and not what it computes: one non-template \texttt{extern "C"} kernel per file, with the template instantiated as the library launches it and helpers such as warp reductions inlined; one operation per line, \texttt{for} written as \texttt{while}, no ternary; warp reductions written as full-mask shuffle steps; a lane-0 guard on the per-warp shared store that every lane makes with the same value; a declared block size where upstream declares none, since the theorem pins the block; dynamic shared buffers as static arrays of a covering size; cache hints and programmatic-dependent-launch calls dropped; a 2-D block flattened; vector accesses kept where the race-freedom lemmas cover them and otherwise written as loops of scalar accesses.

\paragraph{Departures.} A step that changes upstream's semantics is marked as a departure, and the theorem is about the kernel with the departure. They are: 64-bit token and stride indices read as 32-bit (exact at the tested shapes; \texttt{vllm\_silu\_and\_mul}, \texttt{sglang\_rmsnorm}, \texttt{vllm\_rms\_norm}); a runtime alignment test dropped, keeping its aligned path (\texttt{vllm\_rms\_norm}); the \texttt{positions} gather dropped, so the input is the already gathered rows of the cosine and sine cache, and sizes and flags written as literals (\texttt{vllm\_rotary\_embedding}, \texttt{vllm\_fused\_qknorm\_rope}); \texttt{vllm\_fused\_qknorm\_rope} made out of place, writing Q and K to a new buffer; the number of threads written as the host's literal and \texttt{rsqrt.approx} replaced by \texttt{rsqrtf} (\texttt{flashinfer\_rmsnorm}); FP32 input and output made FP16 storage (\texttt{llmc\_softmax\_forward}, \texttt{fam\_flash\_attn}); the running max and sum made per-thread arrays and the output zeroed by the kernel rather than the host, and \texttt{\_\_expf} replaced by \texttt{expf} (\texttt{fam\_flash\_attn}); the int4 dequantization removed, so the weights are FP16, a single split-$K$ iteration, and the block decode's sizes written as literals (\texttt{vllm\_awq\_gemm}, transcribed into PTX from upstream's inline assembly). \texttt{sglang\_fused\_add\_rmsnorm} has no departure and is verified in place, with \texttt{input} and \texttt{residual} as in-out buffers. Per kernel, the rewrite steps and, of them, the departures are: \texttt{vllm\_silu\_and\_mul} 6 and 1, \texttt{sglang\_rmsnorm} 10 and 1, \texttt{sglang\_fused\_add\_rmsnorm} 9 and 0, \texttt{vllm\_fused\_qknorm\_rope} 11 and 3, \texttt{vllm\_rotary\_embedding} 6 and 3, \texttt{vllm\_rms\_norm} 8 and 2, \texttt{flashinfer\_rmsnorm} 12 and 2, \texttt{llmc\_softmax\_forward} 11 and 2, \texttt{fam\_flash\_attn} 9 and 4, \texttt{vllm\_awq\_gemm} 10 and 3.

\paragraph{The llm.c race.} \texttt{softmax\_forward\_kernel7} computes the softmax of each row in three passes: a maximum, a sum of exponentials, and a normalization that writes the row of \texttt{out}. In the third pass a thread's unguarded clamped load \texttt{\_\_half yh = out[a]} with $a = \min(C-1, k)$ reads a cell of \texttt{out} that another thread of the same block stores at \texttt{out[a] = qh} in the same pass, with no barrier between; the loaded value is used only where $k < C$, where the clamped address is the thread's own cell, so on the tested shapes nothing observable depends on the race. What happened in LARK: the translator's race-freedom layer builds its footprints from accesses whose addresses are polynomials in the thread and loop indices, and the clamped address is not one, so it refuses to emit its lemma \texttt{raceFree\_1} for the output store, without locating the race, and the obligation goes to the agent fallback of \Cmain{sec:advanced}, where the race-freedom statement is false. The fallback's loop with GLM~5.3 ended after 10 turns and 9.8M input tokens with a counterexample rather than a proof: thread 0 of block 0 at the load with $k \ge C-1$ reads \texttt{out[50256]}, and thread 80 of the same block, the unique writer of that cell, stores it at the third pass's store; the independent second opinion confirmed the race. A second run with a neutral kernel header found a different witness of the same race (a one-block, 32-thread launch the theorem admits) in 7 turns, so the race is found from the code alone, at about a tenth of what a proof of this kernel costs. The kernel of \Cmain{tab:lib-kernels} is this original text, and its row records the rejection. A one-line rewrite removes the race: upstream reads \texttt{out[row + min(C-1, k)]} unconditionally, and the rewrite reads \texttt{out[row + k]} under the store's bounds guard \texttt{if (k < C)}, so each thread reads only the cell it then writes, with every other line unchanged and the same values on every input; it verified with Opus~5.5 in a separate run.

\ifdefined\appendixonlyflag  
\bibliography{references}
\bibliographystyle{mlsys2025}
\fi
\fi  


\end{document}